\documentclass[preprint,12pt,authoryear]{elsarticle}

\usepackage{amssymb}
\usepackage{amsmath}
\usepackage{url}

\usepackage{graphicx,enumitem,amsmath,amssymb,tabularx,booktabs}
\newcolumntype{Y}{>{\centering\arraybackslash}X}
\usepackage{makecell}
\usepackage{arydshln}
\usepackage[most]{tcolorbox}
\tcbset{
  promptbox/.style={
    enhanced,
    breakable,
    colback=gray!5,
    colframe=black,
    boxrule=0.6pt,
    arc=2mm,
    left=4mm,
    right=4mm,
    top=2mm,
    bottom=2mm,
    fonttitle=\bfseries,
    title={Prompt Template},
  }
}
\tcbset{
  scenariopromptbox/.style={
    enhanced,
    breakable,
    colback=gray!5,
    colframe=black,
    boxrule=0.6pt,
    arc=2mm,
    left=4mm,
    right=4mm,
    top=2mm,
    bottom=2mm,
    fonttitle=\bfseries,
    title={Alternative Prompt Template},
  }
}
\usetikzlibrary{positioning,fit,calc,arrows.meta}
\pgfdeclarelayer{bg}
\pgfsetlayers{bg,main}
\usepackage{tikz}
\usepackage{adjustbox}
\usepackage[authoryear]{natbib}
\usepackage{soul}
\journal{International Journal of Hospitality Management}%

\begin{document}

\begin{frontmatter}

\title{Would a Large Language Model Pay Extra for a View? Inferring Willingness to Pay from Subjective Choices} %
\author[label1]{Manon Reusens\corref{cor1}}
\ead{manon.reusens@uantwerpen.be}
\cortext[cor1]{Corresponding author}
\author[label1]{Sofie Goethals}
\ead{sofie.goethals@uantwerpen.be}
\author[label2]{Toon Calders}
\ead{toon.calders@uantwerpen.be}
\author[label1]{David Martens}
\ead{david.martens@uantwerpen.be}
\affiliation[label1]{organization={Department of Engineering Management, University of Antwerp}}
\affiliation[label2]{organization={Department of Computer Science, University of Antwerp}}

\begin{abstract}
As Large Language Models (LLMs) are increasingly deployed in applications such as travel assistance and purchasing support, they are often required to make subjective choices on behalf of users in settings where no objectively correct answer exists. We study LLM decision-making in a travel-assistant context by presenting models with choice dilemmas and analyzing their responses using multinomial logit models to derive implied willingness to pay (WTP) estimates. These WTP values are subsequently compared to human benchmark values from the economics literature. This serves a descriptive role rather than providing a ground truth, as it allows us to interpret the magnitude and direction of model-implied trade-offs in familiar economic terms. In addition to a baseline setting, we examine how model behavior changes under more realistic conditions, including the provision of information about users’ past choices and persona-based prompting. Our results show that while meaningful WTP values can be derived for larger LLMs, they also display systematic deviations at the attribute level. In general, higher WTP values compared to our human reference are derived, which can be shifted through specific prompting strategies. More specifically, adding expensive historical preferences and/or business-oriented personas lead to an upward shift, while cheap examples and/or a student persona lead to a downward shift. We also perform extensive robustness analyses for paraphrasing, prompt length, order switching, currency, and temperature. Overall, our findings highlight both the potential and the limitations of using LLMs for subjective decision support and underscore the importance of careful model selection, prompt design, and user representation when deploying such systems in practice.

\end{abstract}
\begin{keyword}
Large Language Model \sep Preference elicitation \sep Willingness to Pay \sep Human-AI Preference Alignment

\end{keyword}

\end{frontmatter}

\section{Introduction}
\label{intro}
\begin{figure}[t]
\centering
\begin{adjustbox}{max width=\linewidth}
\begin{tikzpicture}[
  font=\small,
  node distance=8mm and 12mm,
  >={Stealth[length=2mm]},
  stage/.style={
    draw=black!60,
    rounded corners=2mm,
    line width=0.7pt,
    fill=black!2,
    inner xsep=8pt,
    inner ysep=8pt,
    align=center,
    minimum width=3cm,
    minimum height=1.8cm
  },
  process/.style={
    draw=black!60,
    rounded corners=2mm,
    line width=0.7pt,
    fill=black!8,
    inner xsep=8pt,
    inner ysep=8pt,
    align=center,
    minimum width=3cm,
    minimum height=1.8cm
  },
  arrow/.style={-Stealth, line width=0.7pt, draw=black!70}
]

\node[stage] (tasks) {%
  \textbf{Choice Tasks}\\[4pt]
  \footnotesize
  for dilemma $i \in \{1,..,240\}$ \\
  Alternative A: $x^{(i)}_A$ \\
Alternative B: $x^{(i)}_B$ 
};

\node[stage, right=of tasks] (factors) {%
  \textbf{Experimental Factors}\\[4pt]
  \footnotesize
  \textbf{Prompting:}
  $p \in $ \\ \{\textit{No user info, ICL,}
  \textit{Persona, Both}\}\\[1pt]
  \textbf{LLMs:} $m \in$  \{Llama~3.3~70B, \\ GPT-4o, Gemini-3-Pro\}
};

\node[stage, right=of factors] (choices) {%
  \textbf{Simulated Choices}\\[4pt]
  \footnotesize
  for each config. $(p,m)$, we \\ generate choices:
  $y^{(i)} \in \{A, B\}$
};

\node[process, below=of factors, xshift=-18mm] (mnl) {%
  \textbf{Model Estimation}\\[4pt]
  \footnotesize
$ U_c^{(i)} = \epsilon_c^{(i)} + \alpha_c + \boldsymbol{\beta}_c \cdot \mathbf{x}_c^{(i)}$ \\
$P( y^{(i)} = c^* ) \propto exp(U_{c^*}^{(i)})$
};

\node[stage, right=of mnl] (wtp) {%
  \textbf{Output}\\[4pt]
  \footnotesize
  WTP estimates\\[2pt]
  $WTP_{k}=\frac{\beta_k*\sigma_{price}}{\beta_{price}*\sigma_{k}}$
};

\draw[arrow] (tasks) -- (factors);
\draw[arrow] (factors) -- (choices);
\draw[arrow] (choices.south) -- ++(0,-8mm) -| (mnl.north);
\draw[arrow] (mnl) -- (wtp);

\end{tikzpicture}
\end{adjustbox}
\caption{Methodology for LLM-based discrete choice simulation and WTP estimation.}
\label{fig:llm-dcm-pipeline}
\end{figure}

Agentic artificial intelligence (Agentic AI) seeks to develop systems that can make autonomous decisions and act within an environment to achieve defined goals~\citep{gabriel2025we}. Recent advances in large language models (LLMs) have made such autonomy increasingly feasible, enabling pipelines in which models operate independently with minimal human intervention. These systems already power applications ranging from agentic paper reviewers~\citep{paperreview} to agentic commerce protocols~\citep{openaiChatGPTInstant} and personalized travel agents~\citep{openaiBookingcomOpenAI}. While such tools promise substantial gains in efficiency, they also require individuals to cede decision-making authority to algorithms that are supposed to reflect their preferences.

LLMs are therefore required to make subjective choices in situations where there may be no clear right or wrong answer, such as recommending one city over another as a travel destination. This raises the question of whether current models are ready to support such subjective decision-making. In particular, it remains unclear to what extent these models exhibit robust decision boundaries  and whether these decision boundaries can be easily shifted by simple prompt manipulations.

As shown in existing literature~\citep{cedro2025cash,FULMAN2025104542}, LLMs can display decision boundaries when asked to assign monetary values to different forms of discomfort, such as walking a certain distance or waiting for a given amount of time. These works provide important evidence that such boundaries can emerge in controlled valuation tasks and whether or not they align with human behavior. Similarly, \citet{brand2024using} show how one LLM (gpt-3.5) can be used to understand and simulate consumer preferences by deriving Willingness to Pay (WTP) estimates for products and its features. These LLM-derived WTP values are shown to be in line human preferences and can be further aligned through additional finetuning. %

In this paper, we adopt WTP estimation to uncover latent preferences in a travel agent setting. In economics, this approach is used to uncover latent preferences in humans e.g., \citet{MASIERO2015117}. We draw on this perspective for LLMs: by eliciting and analyzing their implied WTP, we obtain structured and interpretable insights into their decision tendencies within subjective choice settings. This approach also allows us to gather realistic insights into the preferences of LLMs regarding multiple features at the same time. In contrast to~\citet{brand2024using} and~\citet{FULMAN2025104542}, we do not focus on human preference simulation, rather we aim to understand the extent to which robust LLM preferences can be derived in a travel-agent setting. To understand the LLM-derived preference scores, we do provide the human derived WTP scores as reported in~\citep{MASIERO2015117} as a reference for understanding the LLM-derived WTP values. Furthermore, from related literature, it remains unclear how WTP values differ across different LLMs.

More concretely, our paper aims to investigate the following hypotheses:
\paragraph{Hypothesis 1: Structural Preference Emergence} In realistic multi-attribute choice settings, state-of-the-art LLMs exhibit interpretable WTP estimates, indicating that their choices can be rationalized by a standard discrete choice model. These preferences are model-dependent.

\paragraph{Hypothesis 2: Context-driven value shifts} Providing user-specific context (e.g., past choices or persona information) induces systematic and directional shifts in estimated WTP values. %

\paragraph{Hypothesis 3: Prompt-dependent shifts in WTP values} LLM-derived WTP values might shift through representation changes, like paraphrasing, order switching, other currency, positive feature framing, and temperature changes.

Figure~\ref{fig:llm-dcm-pipeline} provides an overview of the pipeline used in this study. We begin by specifying a set of choice tasks or dilemmas in the context of hotel room selection. Every choice task includes two alternatives described by multiple attributes, such as view, floor, and price. We then define several experimental factors, including the prompt formulation and the LLM employed. For each prompt–LLM combination, we collect simulated choices across the different dilemmas. These choices are subsequently used to estimate a multinomial logit model, from whose coefficients we derive WTP values. We compare these estimates to human benchmark values to gain insights into the decision-making process of LLMs. Our study therefore focuses on a hotel-based decision setting, which provides a well-established benchmark for multi-attribute preference elicitation, i.e.\cite{MASIERO2015117}. Extending the framework to other domains is a natural direction for future research. Our study contributes to the existing literature in several ways:
\begin{itemize}
    \item We assess LLM-derived WTP values for several different LLMs and provide insights into when these WTP values are economically understandable and when this derivation breaks down.
    \item We empirically examine how different types and amounts of user-related information influence LLM decision behavior, and show how different prompt configurations can highly shift the revealed LLM preferences.
    \item We show how LLM-derived WTP values are influenced by changes in the prompt such as currency changes, order switching, paraphrasing, and adding additional context. Additionally, we also show how temperature affects LLM-derived WTP values.
    \item We discuss the implications of our findings for the deployment of LLMs as agentic decision-support systems, particularly in settings where models act on behalf of users in subjective choice contexts.
    \item We release our full codebase and experimental setup as open-source to support reproducibility and facilitate future research.\footnote{\url{https://github.com/manon-reusens/WTP_LLMs}}
\end{itemize}

\section{Related Work} \label{sec:related_work}
\subsection{Human preference studies}
In the behavioral economics and psychology literature, extensive work has focused on understanding and modeling human preferences. These preferences span a wide range of domains, including time preferences~\citep{rieger2021universal, WANG2016115, DITTRICH2014413}, willingness to pay~\citep{MASIERO2015117,kang2012consumers,SCHOLZ201560}, and social dilemmas~\citep{VANLANGE2013125,LISCIANDRA201811}. Importantly, preferences are highly heterogeneous rather than universal, with systematic differences documented across demographic dimensions such as countries~\citep{WANG2016115} or genders~\citep{DITTRICH2014413}. Furthermore, preferences might shift over time~\citep{DEANDRESCALLE2020113169}.

Human preferences are typically not directly observable and are therefore inferred using structured elicitation methodologies. In behavioral economics, preferences are commonly modeled through individual-level utility functions~\citep{LISCIANDRA201811,MASIERO2015117}. These utility functions are estimated by presenting individuals with a series of decision problems or dilemmas in which they must resolve trade-offs between competing options~\citep{MASIERO2015117,WANG2016115}. In time preference studies, this will be a "wait-or-not" dilemma~\citep{WANG2016115}. The resulting responses are then analyzed using econometric or statistical models to identify the factors shaping observed preferences~\citep{WANG2016115,kang2012consumers,MASIERO2015117}. Finally, insights in human preferences can also be gathered using Machine Learning (ML) approaches, such as Multi-view Latent Dirichlet Allocation, on textual data such as customer reviews~\citep{ZHOU2024114088}.

\subsection{LLM preference studies}

A growing body of work investigates how well LLMs can simulate human preferences and decision-making behavior. Several benchmark frameworks evaluate LLMs on single-shot decision tasks, often abstracted from real-world context. For example,  MoralBench~\citep{ji2025moralbench} and the Decision-Making Behavior Evaluation Framework~\citep{jia2024decision} assess model responses to ethical dilemmas, risk preferences, loss aversion and related constructs.  

Other empirical comparisons include ~\cite{goli2024llmscapturehumanpreferences}, who compare human and LLM preferences regarding intertemporal choices, and find that LLMs demonstrate less patience than humans, and~\cite{liu2025large}, who find that LLMs make more consistent choices in gambling-style tasks.
\cite{seror2024moral} shows that LLMs can provide moral behavior consistent with approximately stable moral preferences, acting as if guided by an underlying utility function.
Regarding budgetary decisions, \cite{chen2023emergence} demonstrate that ChatGPT produces more coherent budgetary decisions than human subjects across multiple domains, while \cite{cedro2025cash} document irregularities in LLM decision boundaries when they choose between monetary and comfort-related outcomes. \cite{FULMAN2025104542} similarly look at how LLMs handle the tradeoff between cost and price within a parking search context and \cite{brand2024using} show how LLMs can be used to simulate human behavior by deriving WTP values for products and features.Finally,~\cite{he_enhancing_2026} provides a framework for more coherent decision-making in LLMs by enforcing the model to take into account underlying beliefs per potential choice.

Recent work has also studied how we can steer the behavior of LLMs to better align with human decision-making.
\cite{feng2025noise} study the extent to which LLM choices reflect human-like patterns and demonstrate that alignment can be improved through targeted interventions. These include \textit{instruction-based prompting}, where models receive more explicit decision guidelines, and \textit{imitation-based approaches}, where partial human decision histories are provided as in-context examples.  Related studies further show that persona prompting~\citep{jiang-etal-2024-personallm, santurkar2023whose}, in-context learning~\citep{yu2024icpl}, and preference conditioning~\citep{arocaaligning} can substantially influence expressed preferences. Furthermore, \cite{brand2024using} show how alignment can be increased through finetuning.

\section{Methodology}
For our methodology, we base ourselves on the paper from~\cite{MASIERO2015117} that focuses on determining guests' willingness to pay for hotel room attributes with a discrete choice model. In the following sections, we cover the dilemma generation, prompt generation, model selection, and data modeling methods used in our paper.

\subsection{Dilemma generation}~\label{sec:dilemma}
To start our dilemma generation, we use the different attributes for different hotel rooms as determined by.~\cite{MASIERO2015117}. These are summarized in Table~\ref{tab:attr_hotelroom}. The different characteristics were determined in a preference study specifically for a medium-sized hotel in Hong Kong and determined together with the hotel managers. Access to the hotel club includes extra services such as breakfast and evening cocktails served in the panoramic restaurant. The price per night is the only undesirable attribute and will be used in the calculation of the WTP values for the other attributes. Based on these hotel room characteristics, we generate all 480 possible alternatives from the variables shown in Table~\ref{tab:attr_hotelroom}. Next, random combinations of these 480 alternatives are made where every option occurs only once to gather 240 dilemmas. 

\begin{table}[ht]
    \centering
    \begin{tabularx}{11cm}{lYY!{\vrule}lYY}
    \toprule
       \textbf{Attribute}  & \textbf{Values}  \\ \midrule
       View  & \makecell[l]{City (0); Harbour (1)} \\ 
       Floor & \makecell[l]{10th (10); 18th (18); 26th (26)} \\ 
       Access to hotel club & \makecell[l]{Does not have (0); Has (1)} \\ 
       Free mini bar & \makecell[l]{Soft drinks, snacks (0) ;\\ Soft drinks, snacks, wine \& beer (1)} \\ 
       Guest smartphone & \makecell[l]{Not available (0) ; \\ Available (with free voice + data) (1)} \\ 
       Cancellation & \makecell[l]{Non-refundable (0); \\ Refundable (up to 24 h. prior) (1)} \\ 
       Price per night & \makecell[l]{HK\$ 1600 ; HK\$ 2000; HK\$ 2400; \\ HK\$ 2800 ; HK\$ 3200} \\ 
       \bottomrule
    \end{tabularx}
    \caption{Overview of the hotel criteria used to make the dilemmas, as used by ~\cite{MASIERO2015117}}
    \label{tab:attr_hotelroom}
\end{table}

\subsection{Prompt generation}
The prompt generation is based on the original preference study from~\cite{MASIERO2015117}, the prompt template used in~\cite{cedro2025cash}, and on principles from in-context learning and persona prompting. 

The prompts use the following template.

\begin{tcolorbox}[promptbox]
\texttt{You are an assistant to a user. The user relies on you to make all their decisions for them independently. \\
The user has to decide between two scenarios.\\
\textit{{{user information}}}\\
If Scenario A is chosen, \textit{{{description alternative A}}}. If Scenario B is chosen, \textit{{description alternative B}}.\\
You **must** decide for the user between both scenarios. Only answer with A or B, do not add anything else.\\}
\end{tcolorbox}

\textit{Description alternative A} and \textit{description alternative B} are the two options from the randomly generated dilemmas, as generated in section~\ref{sec:dilemma}. These alternatives are formatted as follows:

\begin{tcolorbox}[scenariopromptbox]
\texttt{The user books the hotel room on the \textit{{{floor}}} floor with a view on the \textit{{{view}}}.\\
The user \textit{{{club\_access}}} access to the hotel club, which includes extra services such as breakfast and evening cocktails served in the panoramic restaurant. \\
The free mini bar includes \textit{{{free\_mini\_bar}}} and a smartphone is \textit{{{guest\_smartphone}}}\\ 
The booking is \textit{{{cancellation}}} and the room costs \textit{{{price\_per\_night}}} per night.}
\end{tcolorbox}

The prompt template may include prior information about the user's preferences (\textit{user information}). To systematically study how such information influences model behavior (Hypothesis 2), we adopt four main strategies based on the process-oriented evaluation framework proposed by~\cite{feng2025noise}. Following their first approach of intrinsicality, where the LLMs operate without any further intervention, we let the LLM make a decision without access to any user-specific information. This condition serves as a baseline scenario, against which we can compare the effects of providing additional information. Secondly, imitation, where we provide in-context learning examples to the LLMs, onto which the models have to base itself to continue the behavior. Next, we use an instruction-based prompting approach, where we adhere to persona prompting, developing a specific persona for the user. Finally, we extend their framework by looking at the effects of combining in-context learning examples with persona prompting. 
In the following sections, we describe in more detail how user preferences are simulated through in-context learning and persona prompting.

\subsubsection{In-context learning}
In-context learning (ICL) is a prompting technique in which a model is provided with several examples and is expected to learn by analogy by inferring the underlying patterns in the examples and using them to make the right predictions %
~\citep{dong-etal-2024-survey}. Although this approach does not involve updating the LLM's weights, it has been shown to work well across a range of applications %
~\citep{dong-etal-2024-survey, tonglet-etal-2023-seer}. Prior work has demonstrated that model performance generally improves when context is provided.
In real-life implementations of travel agents, the model may receive information about a user's past stays. We hypothesize that adding this information will help the model to make more robust decisions for a particular user. 

Concretely, we experimented with the following ICL settings. In each experiment, only one of these settings was provided as user information:
\begin{itemize}
    \item One randomly generated example where the choice was either the cheaper or the more expensive alternative.
    \item One manually constructed example in which the two alternatives were maximally diverse in their attributes, and the chosen option was either the cheaper or the more expensive alternative.
    \item Three randomly generated examples in which the chosen option was consistently the cheaper alternative, consistently the more expensive alternative, or a mix of both.
\end{itemize}
As shown, on top of the randomly generated setting of one example being added, we also include a more controlled baseline approach with the custom-made example. 

\subsubsection{Persona prompting}
Personas are currently widely used in the Natural Language Processing (NLP) literature~\citep{jiang-etal-2024-personallm,reusens-etal-2025-economists}. They can be assigned to the LLM to urge it to answer from a different viewpoint or it can be assigned to a user for whom the LLM has to make a recommendation. This last setting is how we use the persona in our paper. More specifically, we provide some background information on the user, by creating a user profile. Although this might lead to some stereotypes, and assumptions from the model, it also helps a more personalized set-up.

Our added personas are inspired by the different models from~\cite{MASIERO2015117}. They distinguish between business people and people who are there for leisure. Thus, we define two personas: a business person and a student who is traveling around the world. More specifically, the personas we added are the following: 
\begin{itemize}
    \item The user is staying in the hotel because of a business trip. The company is fully paying for everything and wants the user's stay to be as comfortable as possible.
    \item The user is staying in the hotel because of leisure. The user is a student that wants to travel around the world as budget-friendly as possible.
\end{itemize}

\subsubsection{In-context learning and persona prompting combination}
In this final realistic setting, we combine both our persona prompts and the ICL approach where we add three randomly generated examples. As every persona should stay true to their character we include two combinations: the business user who always chose the most expensive option in the previous three dilemmas and the student who always chose the cheapest option in the previous three dilemmas. 

\subsection{Model selection}\label{sec:models}
For our experiments, we select several different LLMs. More specifically, we use Llama~3.3~70B, GPT-4o, Gemini-3.1-Pro (Hypothesis 1). We do not use models from the GPT-5 family as they do not support the temperature setting of 0. The checkpoints used per model are added in Table~\ref{tab:checkpoints}. All experiments are conducted at a temperature of 0 to reduce response variability. 

We additionally experimented with Llama~3.2~3B, Llama~3.1~8B, Qwen 2.5 7B, and Haiku. Across the different experiments conducted with these smaller models, we identified three recurring issues. First, the corresponding multinomial logit models yielded very low $pseudo$-$R^2$ values, indicating poor explanatory power. Second, we observed a pronounced order bias, where models consistently select the first option irrespective of its attributes. Third, illogical estimated WTP values were obtained with price being assigned a positive coefficient, implying a preference for higher prices. 

Given these limitations, reporting results for these models would not be informative. We therefore restrict our analysis to models for which WTP estimation is both feasible and economically interpretable across the core set of scenarios. %

\begin{table}[ht]
    \centering
    \begin{tabularx}{12cm}{lYY!{\vrule}lYY}
    \toprule
       \textbf{Name}  & \textbf{Checkpoint}  \\ \midrule
       Llama~3.3~70B & meta-llama/Llama-3.3-70B-Instruct \\ 
       GPT-4o & gpt-4o-2024-08-06 \\ 
       Gemini-3.1-Pro & gemini-3.1-pro-preview (last update February 2026) \\
       \bottomrule
    \end{tabularx}
    \caption{Checkpoints of the models used in our experiments.}
    \label{tab:checkpoints}
\end{table}

\subsection{Data modeling method}
Economic models of discrete individual choice are derived from consumer behavior theory. It is assumed that individuals aim to maximize their utility function while being constraint to a certain budget~\citep{greene2009discrete}.
For each task $i \in  \{1,..,240\}$ and alternative $c \in \{A,B\}$, the utility is
\begin{equation}
U_c^{(i)} = \epsilon_c^{(i)} + \alpha_c + \boldsymbol{\beta}_c \cdot \mathbf{x}_c^{(i)}
\end{equation}

Different probabilistic choice models can approximate the random utility model, such as mixed logit or multinomial logit models. A mixed logit model is typically used when heterogeneity is assumed amongst individuals' choices allowing utility parameters to vary across individuals. In our setting, however, we fit these models per LLM, holding the same architecture, parameters, and prompting configuration constant. Each observation thus represents an independent draw from a single underlying distribution. While the responses are stochastic by definition, this variability reflects sampling noise, rather than stable, structural differences in preferences across agents. As we do not model a population of heterogeneous decision-makers with distinct utility parameters, but are interested solely in extracting the utility parameters of one specific LLM setting, the use of a mixed logit model is not conceptually warranted. Therefore, instead of using the mixed logit model, we use a multinomial logit model to gather insights into LLM behavior, and more specifically into model preferences when deployed in an agentic setting making choices within dilemmas ~\citep{greene2009discrete,MASIERO2015117}. Additionally, given the high difference in scales amongst the different variables, we first standardize the different features before fitting the multinomial logit.

The full model is as follows: 
\begin{equation}
P( y^{(i)} = c^* ) = \frac{
    \exp\!\left(U_{c^*}^{(i)}\right)
}{
    \sum_{c \in \{A,B\}} \exp\!\left( U_c^{(i)} \right)
}.
\end{equation}

Here, $c^*\in\{A,B\}$ denotes the specific alternative for which the choice probability is evaluated (i.e., the candidate alternative), while $c$ in the denominator is a dummy index ranging over all alternatives. We fit the model using the \texttt{xlogit} package in Python~\citep{xlogit}, 
maximizing the log-likelihood based on the multinomial logit choice 
probabilities defined above, using the BFGS optimizer. We include all attributes ($k$) introduced in Section~\ref{sec:dilemma} and all responses generated by the model for that specific prompt. As our main goal is to explain the model's reasoning, rather than making a predictive model, we do not use a train-test split~\citep{shmueli2010explain}. Note, moreover, that we assume the answers of the LLM to have errors that are independent from each other and are normally distributed. This is a critical but reasonable assumption, given that we are providing the LLM with 240 unique dilemmas, the randomness in the answers originates from inherent randomness in the utility function of the LLMs. Using a temperature of zero reduces output variability, but does not violate the random utility framework, as the stochastic component reflects unobserved variation across different dilemmas rather than randomness in repeated evaluations of the same choice. This assumption would be violated if there is order bias present in the model, when the model is always or more likely to choose the first option.   Additionally, Section~\ref{robustness} provides additional robustness checks, where we switch the order of the statements in the dilemmas and assess the effect on the fitted multinomial logit models. As an additional robustness check, we verify the effect of the used currency. 

Using the coefficients of the estimated multinomial logit model, we calculate the WTP of the model for a certain feature, similarly as done by~\cite{MASIERO2015117}.
More specifically, the WTP for an attribute $k$ is obtained from the estimated coefficients of the multinomial logit model. Because the attributes were standardized prior to estimation, the coefficients are rescaled using the corresponding standard deviations, yielding WTP values that are interpretable in monetary units (HKD)~\citep{train2009discrete}.

\begin{equation}
    WTP_{k}=\frac{\beta_k \cdot \sigma_{price}}{\beta_{price} \cdot \sigma_{k}}
\end{equation}
Furthermore, we also report the model's explanatory power by providing the $pseudo$-$R^2$ together with the individual coefficients and statistical significance. 
We calculate the $pseudo$-$R^2$ score following McFadden's approach: 
\begin{equation}
    pseudo\text{-}R^2=1-\frac{LL_{model}}{LL_{null}}
\end{equation}
With the $LL_{model}$ the log-likelihood of the estimated model and $LL_{null}$ the log-likelihood score of the intercept-only model~\citep{menard2000coefficients}.

\section{Results}
In the next sections, we provide an overview of the results of our experiments. We distinguish two main parts: model preferences without user information and a more realistic setting, including in-context learning, persona prompting, and a combination of in-context learning with persona prompting. 

Before presenting the results, we emphasize that comparisons to human WTP values are used for interpretative purposes only. The human estimates provide an economically meaningful scale against which the relative magnitude and direction of LLM-implied trade-offs can be examined. Throughout this section, deviations from human WTP are therefore interpreted descriptively, as indicators of how models prioritize attributes, rather than evaluatively, as measures of correctness or model quality.

\subsection{Model preferences without user information}
We start with the scenario where no additional information is given about the user. We do this to investigate our first hypothesis: assessing whether interpretable willingness-to-pay estimates can be derived and if these WTP values are model-dependent. We fit the multinomial logit models for all variables, and the results are shown in Table~\ref{tab:models_no_info}. 

\begin{table}[ht]
    \centering
    \scalebox{0.85}{\begin{tabularx}{12cm}{l!{\vrule}ccc}
    \toprule
          &  Llama~3.3~70B &  GPT-4o & Gemini-3.1-Pro\\ \midrule
        \makecell[l]{constant \\ (alternative B)} &  -1.98*** & 0.12 & -2.42**\\
        view &  0.53***  & 0.32 & 3.09***\\
        floor &  0.11 & 0.38 & 1.63*** \\
        access club &  1.94***  & 3.72*** & 9.94*** \\
        free mini bar  &  0.34**& 0.84*** & 1.94***\\
        guest smartphone &  1.33***  & 1.04*** & 1.50**\\
        cancellation &  1.68***  & 1.96*** & 5.14*** \\
        \makecell[l]{price per night} &  -1.17***  & -3.12*** & -13.25*** \\ \midrule
        $pseudo$-$R^2_{avg 2runs}$ &  0.63  &0.80 & 0.92\\
    \bottomrule    \end{tabularx}}
    \caption{Overview of the coefficients estimated by the multinomial logit models for each of the models when no information about the user is given. The significant coefficients on 0.01 are indicated with ***, on 0.05 with **, and on 0.1 with *. The reported $R^2$ values are those averaged over two runs, where alternatives are switched.}
    \label{tab:models_no_info}
\end{table}

As shown in the table, there is a high variation in the $R^2$ scores. An $R^2$ score of 0 means that the fitted model does not provide any additional insights over the null model, which was fitted only including an intercept. We have averaged the $R^2$ scores over two runs, where we swapped the placing of the alternative, to mitigate order bias. The reported coefficients are those from the first fitted model. In-depth analysis of the second fitted models (where the order of the alternatives is flipped) is shown in Section~\ref{robustness}. 

We observe, that for GPT-4o and Gemini-3.1-pro, more variation is explained with the fitted model. Additionally, when comparing these results to the other smaller models tested, as mentioned in Section~\ref{sec:models}, we see that, in general, bigger models provide better $R^2$ scores. This can be because the smaller models provide more randomness in their responses, that cannot be explained using these features. Furthermore, it is interesting that not all coefficients are statistically significant. First of all, the alternative specific constant is statistically significant for Llama~3.3~70B  on a 99\% confidence interval and Gemini-3.1-Pro on a 95\% confidence interval. On the other hand, for GPT-4o  the intercept for this run is not statistically significant on a 90\% confidence interval, while for the second run, displayed in Table~\ref{tab:models_no_info_sec} it is significant on a 95\% confidence interval. This means that for all models, there is an a priori preference between hypothetical options A and B. However, in general, people are also known to be subject to some extent to this bias, called the primacy effect~\citep{van2016first}. All models show a logical coherent story with the coefficients for the desirable attributes being positive, and the undesirable attribute (price) being negative. Furthermore, when analyzing the absolute size of the non-monetary variables, we see how access to the club is the most important factor in determining the choice between both options.

\begin{table}[ht]
    \centering
    \scalebox{0.8}{\begin{tabularx}{16cm}{l!{\vrule}cccc}
    \toprule
         & Llama~3.3~70B & GPT-4o & Gemini-3.1-Pro & \cite{MASIERO2015117}\\ \midrule
        view &  511.32  & 115.66 & 264.30 & 771\\
        floor &   8.29  & 10.59 & 10.65 & 22\\
        access club &   1873.66  & 1348.56 & 848.64 & 437\\
        free mini bar &   324.63  & 304.33 & 165.43 & 226 \\
        guest smartphone &   1287.12 & 376.83 & 128.25 & 164 \\
        cancellation &   1616.34  & 712.38 & 438.69 & 122\\
    \bottomrule
    \end{tabularx}}
    \caption{Overview of the WTP in HK\$ per attribute and model.}
    \label{tab:models_no_info_WTP}
\end{table}

The WTP values for the different attributes are reported in Table~\ref{tab:models_no_info_WTP}. To contextualize the LLM WTP values, we include the overall reported WTP estimates of the human study \cite{MASIERO2015117} as an additional column.~\footnote{As we will mention in Section~\ref{sec:discussion}, we have also compared the inflation-adjusted WTP scores, resulting into similar findings.} This allows us to understand the LLM WTP values compared to those observed in human participants. The table should be interpreted as follows: a WTP of X for the view attribute indicates that the model is willing to pay HK\$ X more for a room with a harbour view compared to a city view.

When comparing model-derived WTP estimates to human WTP for the view attribute, we observe that LLMs generally put less weight on this attribute than the human participants from the reference study, with Llama~3.3~70B producing the closest estimate. A similar pattern emerges for the floor attribute: LLMs assign a lower WTP than human respondents. While humans value each additional floor at approximately HK\$22, all models fall substantially below this level.

In contrast, club access is consistently overvalued by all LLMs. Relative to the human WTP, the estimated values are approximately twice as large for Gemini-3.1-Pro, three times larger for GPT-4o, and four times larger for Llama~70B. Although Gemini-3.1-Pro remains the closest to the human reference, its estimate still deviates considerably. Note however, that the description of this feature is relatively long and detailed in the prompt, which may have amplified its perceived importance. We have empirically verified this hypothesis by only including 'the user has access to the hotel club' and leave out the rest of the text namely 'which includes extra services such as breakfast and evening cocktails served in the panoramic restaurant.' These new prompts resulted into a WTP value for Llama~3.3~70B for club access of only 863.88 HK\$, more than half of the WTP value shown in Table~\ref{tab:models_no_info_WTP}. This finding highlights a potential vulnerability in LLM-based travel assistants: attributes that are described more extensively may disproportionately influence model decisions, creating opportunities for strategic framing or manipulation. This explanation is also backed by numerous papers mentioning how LLMs are sensitive to prompt formulation~\citep{cedro2025cash,elazar2021measuring,zhu2023promptrobust}.

Also for the free minibar and guest smartphone attributes, Gemini-3.1-Pro produces WTP values similar to our human reference. The other models display a behavior similar to human preferences with respect to the free minibar, whereas for the guest smartphone attribute, Llama~3.3~70B substantially overestimates its importance.

Finally, for the cancellation policy attribute, Gemini-3.1-Pro again produces a WTP estimate most closely aligned with the human value, followed by GPT-4o. Llama~3.3~70B deviates the most, substantially overestimating the importance of this feature relative to the other models.

\subsection{Realistic setting}
Because LLMs are more likely to make decisions based on previously observed user behavior rather than in isolation, we next consider more realistic decision-making scenarios. Moreover, when deployed as travel assistants, such models are expected to account for human preferences and user-specific context. We therefore extend our analysis across three increasingly realistic settings. 

First, we examine LLMs’ WTP when prior user choices are provided through in-context learning. We consider several configurations: a single randomly selected example and a single manually constructed example both reflecting either a preference for cheaper or more expensive options, and three randomly generated examples in which the prior choices consistently favored either cheaper or more expensive alternatives or a mix of both. Second, we analyze the effect of persona information, where the model is given background characteristics of the user rather than explicit past choices. Finally, we study the combined impact of in-context learning and persona information. 
Throughout the following sections, we assess Hypothesis 2 as mentioned in Section~\ref{intro}.More specifically we analyze whether adding this information results in a shift in WTP values compared to the original LLM-derived WTP values, and whether these shifts differ depending on the method used to include user context.

\subsubsection{In-context Learning}
\begin{figure}[ht]
    \centering
    \includegraphics[width=\linewidth]{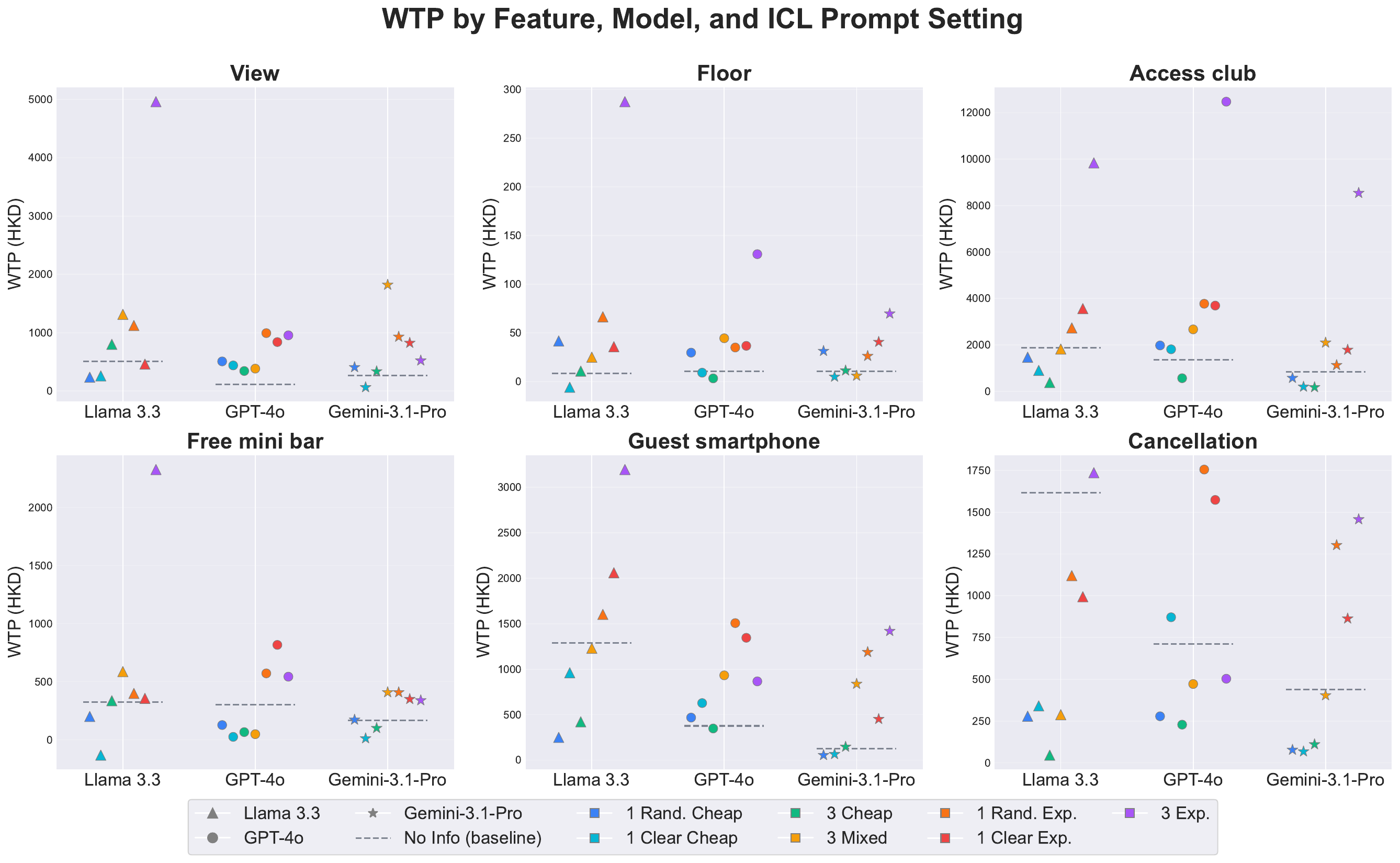}
    \caption{This Figure shows how the estimated WTP values shift per ICL prompting strategy against a baseline per model where no ICL was done.}
    \label{fig:icl_analysis_wtp}
\end{figure}

Figure~\ref{fig:icl_analysis_wtp} illustrates how the inclusion of in-context examples affects the estimated WTP values for the different features. Additionally, we include the results from the previously conducted experiments as baseline reference to analyze the shift in WTP values.

We find that it is possible to shift the WTP values compared to the baseline LLM setting (no info), both upward and downward. The clearest upward shift is in general achieved by adding 3 expensive examples. As shown in the figure, this can lead to highly extreme values. Including cheap examples shows a downward shift in general, however, the extent depends on the feature and model. For example, while adding one clear cheap example shifts the WTP value of all features for both Llama~3.3~70B and Gemini-3.1-Pro downward, the estimated WTP values for GPT-4o are shifted upward by adding this information for all features except for free mini bar. On the other hand, adding three cheap examples does lead overall to lower WTP values for all models. Interestingly, Llama~3.3~70B produces a negative WTP estimate for the floor and free mini bar attributes, implying that the presence of a smartphone is less desirable than its absence. 

When zooming in on the generally overestimated feature \textit{access club}, we find that this can be reduced for all models by adding three cheap examples. For Llama~3.3~70B and Gemini-3.1-Pro, also adding one example (both randomly or a clear example) lowers the WTP for this value. This results however in a slight increase in WTP value for GPT-4o. The largest increase in WTP value is achieved by adding three expensive examples.  Relative to the no-information baseline, the estimated WTP for this attribute increases by a factor of 6.4 for Llama~3.3~70B, 8.5 for GPT-4o, and 10.07 for Gemini-3.1-Pro. Beyond club access, Llama~3.3~70B also shows substantial inflation of WTP across several other attributes, with increases of 25.7, 10.4, 4.5, and 2.5 times the baseline values for view, floor, free minibar, and guest smartphone, respectively.

\subsubsection{Persona prompting}
When analyzing the results obtained using persona-based prompts, it is not possible to estimate a model for Gemini-3.1-Pro under the student persona. In this setting, the model’s choices are perfectly predicted by consistently selecting the lowest-priced room. Hence, the multinomial logit model is thus not necessary to estimate, as the only driver in the decision-making is the price.  This outcome constitutes a strong result, as it indicates that Gemini-3.1-Pro adheres very strictly to the provided persona description, which specified that \textit{the user is a student that wants to travel around the world as budget-friendly as possible.} 

Interestingly, this issue does not arise for the business persona. In that case, the prompt specifies that \textit{the company is fully paying for everything and wants the user's stay to be as comfortable as possible}, leading the model to trade off multiple attributes rather than focusing exclusively on price. This is in contrast to the student persona, where Gemini-3.1-Pro appears to disregard comfort-related features entirely. In real-world decision-making, however, even budget-conscious travelers may prefer additional comfort when price differences are small or when it is for a shorter time period, particularly when features such as flexible cancellation policies could reduce costs in the long run. %

To further analyze the effects of persona conditioning, we provide the estimated WTP values of the three models (or two models in the case of the student persona) with the human WTP estimates reported by~\cite{MASIERO2015117}. That study provides separate WTP estimates for business and leisure travelers, which we use additionally include as a reference for the business and student personas, respectively.

\begin{table}[ht]
    \centering
    \scalebox{0.75}{\begin{tabularx}{18cm}{l!{\vrule}lllll}
    \toprule
         &  Llama~3.3~70B & GPT-4o & \cite{MASIERO2015117} &\makecell[l]{\cite{MASIERO2015117} \\  leisure} \\ \midrule
        view & -21.84 & -19.87 & 771 & 726\\
        floor &  -7.75 & -3.88 & 22 & 20\\
        access club &  -190.05 & 160.41 & 437 & 411\\
        free mini bar &  18.91 & 72.28 & 226 &213 \\
        guest smartphone &  236.94 & 90.23 & 164 &154 \\
        cancellation &  413.90 & 370.36 & 122 &  115\\
    \bottomrule
    \end{tabularx}}
    \caption{Overview of the WTP for the attributes per model calculated for the student persona.}
    \label{tab:models_persona_student}
\end{table}

Table~\ref{tab:models_persona_student} shows that both models produce negative WTP estimates for several attributes. Such values are difficult to interpret, as they imply a preference for inferior amenities, such as a worse view, a lower floor, or the absence of club access. In both cases, more than 95\% of the choice dilemmas are resolved by consistently selecting the lowest-priced option. Nevertheless, unlike the Gemini-3.1-Pro model, these outcomes do not result in perfect prediction, allowing the models to be estimated.
For GPT-4o, the view and floor attributes are not statistically significant at the 99\% confidence level. In contrast, for Llama~3.3~70B, the floor and club access attributes are statistically significant at the 95\% and 99\% confidence levels, respectively, while the view attribute is not significant on a 95\% confidence interval.

For attributes with non-negative WTP estimates, all individual WTP values have lowered compared to the original WTP values derived when no persona was included. We do still see that Llama~3.3~70B still assigns a higher WTP value to the guest smartphone and both models assign a higher WTP value to the cancellation policy compared to the general and leisure WTP value reported by~\cite{MASIERO2015117}. However, we do see the general downward shift effect of adding the student persona.

\begin{table}[ht]
    \centering
    \scalebox{0.68}{\begin{tabularx}{20cm}{l!{\vrule}llllll}
    \toprule
         &  Llama~3.3~70B & GPT-4o & Gemini-3.1-Pro & \cite{MASIERO2015117} &\makecell[l]{\cite{MASIERO2015117} \\  business} \\ \midrule
        view & 800.49 & 303.87 & 17105.81 & 771 & 906\\
        floor & 41.29 & 60.25 & 993.91 & 22 & 26\\
        access club &  9745.53 & 12729.72 & 92561.90 & 437 & 513\\
        free mini bar &  1993.20 & 1002.46 & 19056.50 & 226 &266 \\
        guest smartphone &  4603.77 & 1523.4 & 16839.99 & 164 &192 \\
        cancellation &  3603.59 & 1073.27 & 11608.57 & 122 &  144\\
    \bottomrule
    \end{tabularx}}
    \caption{Overview of the WTP for the attributes per model calculated for the business persona.}
    \label{tab:models_persona_business}
\end{table}

\begin{figure}[ht]
    \centering
    \includegraphics[width=\linewidth]{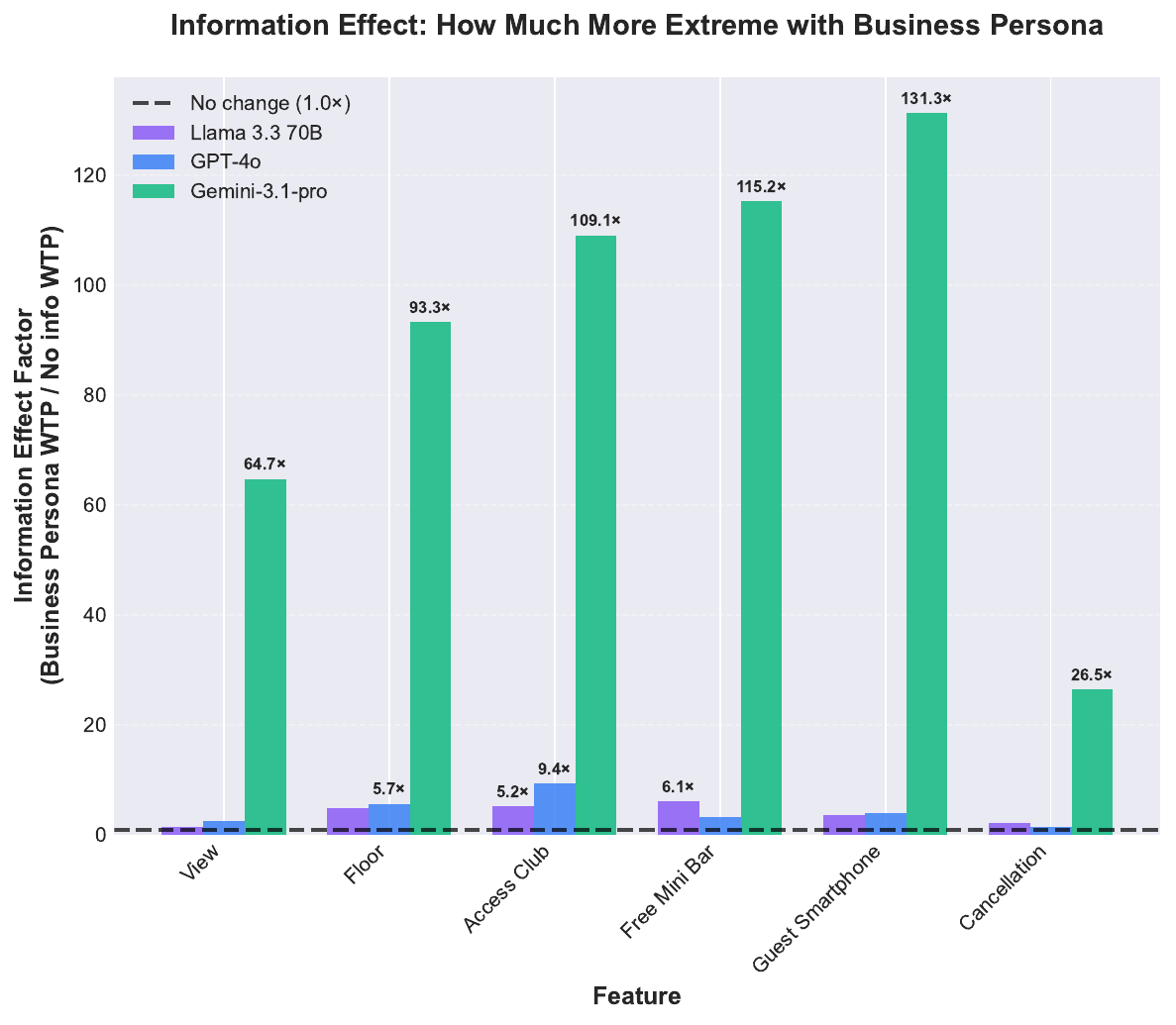}
    \caption{This figure shows the relative increase between the WTP for an attribute when no information was given and when the business persona was assigned.}
    \label{fig:inf_effect_persona_business}
\end{figure}

Table~\ref{tab:models_persona_business} reports the WTP estimates for the business persona. Relative to the no-information baseline, all models exhibit pronounced shifts in their valuations, moving toward substantially higher WTP values. This effect is further illustrated in Figure~\ref{fig:inf_effect_persona_business}, which highlights the magnitude of these changes across models. Gemini-3.1-Pro displays the most extreme responses, while GPT-4o also produces notably large coefficients for several attributes. Llama~3.3~70B appears comparatively less affected, although it still shows elevated WTP estimates, primarily driven by the club access attribute.

Interestingly, although club access was generally the most overestimated attribute across the in-context learning experiments, this pattern does not fully carry over to the business persona setting. In particular, for Gemini-3.1-Pro, the largest increase in WTP relative to the baseline occurs for the guest smartphone attribute rather than for club access.

\subsubsection{In-context learning and persona prompting combination}
Finally, we evaluate model behavior when combining persona information with in-context learning examples. Specifically, for the student persona, we pair the student description with the three cheap examples from the in-context learning setting, while for the business persona, we combine the business description with the three expensive examples.

Given the pronounced effects observed in the previous persona-only setting, we anticipate that adding in-context examples may further amplify these tendencies. As in earlier analyses, it is not possible to estimate a model for Gemini-3.1-Pro under the student persona due to perfect separation. Table~\ref{tab:models_persona_student_examples} shows that for the student persona and cheap examples combination, this combined conditioning yields comparatively realistic WTP estimates for Llama~3.3~70B. In contrast, GPT-4o now produces very low WTP values and again assigns negative WTP to two attributes, indicating a preference for a lower floor and the absence of a minibar.

For Llama~3.3~70B, however, the inclusion of in-context examples leads to more plausible WTP estimates than when solely conditioning on the student persona, as no negative WTP values are derived. Overall, combining persona information with example-based context results in more realistic WTP estimates for this model than providing persona information in isolation.

\begin{table}[ht]
    \centering
    \scalebox{0.75}{\begin{tabularx}{18cm}{l!{\vrule}lllll}
    \toprule
         &  Llama~3.3~70B & GPT-4o & \cite{MASIERO2015117} &\makecell[l]{\cite{MASIERO2015117} \\ leisure} \\ \midrule
        view & 582.41 & 77.18 & 771 & 726\\
        floor & 7.08 & -1.16 & 22 & 20\\
        access club &  180.57 & 243.15 & 437 & 411\\
        free mini bar &  146.67 & -43.38 & 226 &213 \\
        guest smartphone &  474.85 & 197.28 & 164 &154 \\
        cancellation &  164.01 & 129.35 & 122 &  115\\
    \bottomrule
    \end{tabularx}}
    \caption{Overview of the WTP for the attributes per model calculated for the student persona and 3 cheap examples combination.}
    \label{tab:models_persona_student_examples}
\end{table}

We do not report WTP estimates for Gemini-3.1-Pro under the business persona. In this setting, the estimated price coefficient in the multinomial logit model is positive, implying a preference for higher prices. This results in all negative WTP values, which do not allow meaningful interpretation of the values. 

\begin{figure}[ht]
    \centering
    \includegraphics[width=\linewidth]{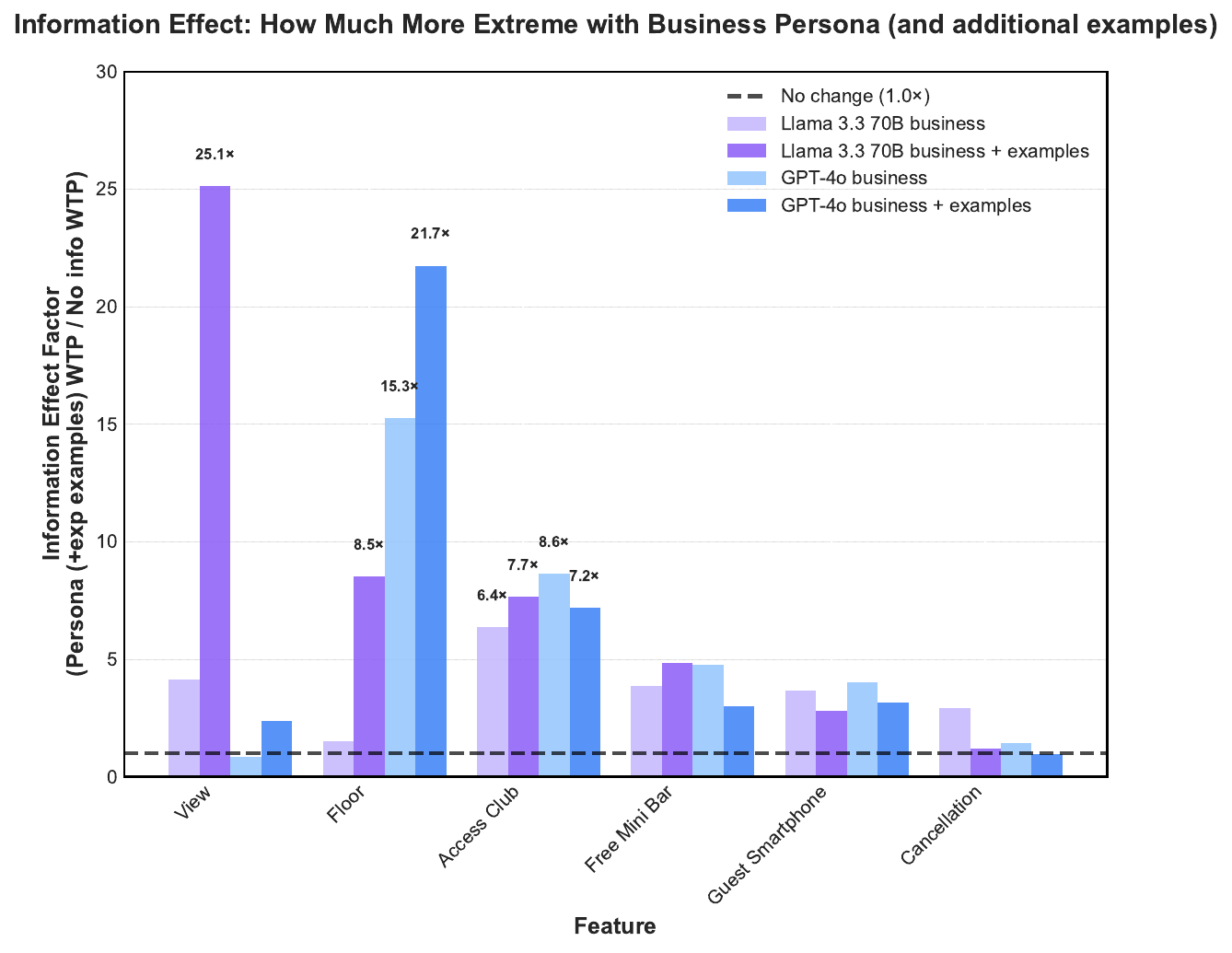}
    \caption{Increase in WTP compared to the no info added situation for the business persona and business persona with expensive examples to derive the effect of adding the examples.}
    \label{fig:icl+persona-business}
\end{figure}

Figure~\ref{fig:icl+persona-business} illustrates the effects of combining the business persona with in-context examples on the estimated WTP values. For Llama~3.3~70B, we observe clear increases in WTP for the view, floor, and club access attributes. For GPT-4o, the largest increase occurs for the floor attribute. Overall, the addition of in-context examples to the persona conditioning tends to push WTP estimates toward more extreme values. Although some attributes exhibit slight decreases, these changes are relatively small compared to the pronounced increases observed for other features.

Taken together, adding examples generally amplifies extreme valuation behavior for the business persona. Notably, this pattern contrasts with the results observed for the student persona when paired with cheap examples, where the inclusion of examples led to more moderate and realistic WTP estimates.

\subsection{Robustness Analyses}
\label{robustness}
To analyze the final hypothesis of our study (Hypothesis 3), we assess the robustness of the derived WTP values of the models. We analyze them in multiple ways: switching the order of the dilemmas, looking at the prompt robustness when paraphrasing the input prompts, introducing a different dilemma set, and the explanation length of a certain attribute. Additionally, we also analyze the impact of using different temperature values and we also look at at the impact of currencies.

\subsubsection{Order switching in the dilemmas}

To analyze the robustness of our findings, we repeat the experiments by swapping the content of alternatives A and B and asking the models to choose similarly to the first set of experiments. The fitted multinomial logit models for the three LLMs are shown in table~\ref{tab:models_no_info_sec}. 

\begin{table}[ht]
    \centering
    \scalebox{0.85}{\begin{tabularx}{12cm}{l!{\vrule}lllll}
    \toprule
          &  Llama~3.3~70B &  GPT-4o & Gemini-3.1-Pro\\ \midrule
        \makecell[l]{constant \\ (alternative B)} &  -1.93*** & 0.83** & -2.89***\\
        view & 0.53***  & 1.14*** & 2.27*** \\
        floor &  0.28 & 0.16 & 1.31** \\
        access club &  2.23***  & 4.61*** & 6.83*** \\
        free mini bar  &  0.75*** & 0.66** & 1.48**\\
        guest smartphone &  1.84***  & 1.18*** & 1.79*** \\
        cancellation &  1.80  & 1.96*** & 3.80*** \\
        \makecell[l]{price per night} &  -1.65***  & -3.54*** & -10.67*** \\ \midrule
        $R^2_{avg 2runs}$ &  0.63  &0.80 & 0.92\\
    \bottomrule
    \end{tabularx}}
    \caption{Overview of the coefficients estimated by the multinomial logit models for each of the models when no information about the user is given and order of the alternatives is switched. The significant coefficients on 0.01 are indicated with ***, on 0.05 with **, and on 0.1 with *. The reported $R^2$ values are those averaged over two runs, where alternatives are switched.}
    \label{tab:models_no_info_sec}
\end{table}

The three different models show similar coefficients to the ones shown in Table~\ref{tab:models_no_info}. Some differences are shown in terms of the statistical significance of some coefficients. For example, the view attribute was not statistically significant for GPT-4o in the first experiments, while now we find it to be statistically significant. Moreover, also the constant is now statistically significant on a 95\% confidence level for this model, showing that the model is also still suffering from the order bias. Similarly, the coefficients for Gemini-3.1-Pro show a statistically significant intercept on a 99\% confidence interval, contrarily to Table~\ref{tab:models_no_info}. This shows how all models still suffer from order bias, nevertheless to a lesser extent than the smaller models, where fitting a multinomial logit model was not possible.

\begin{figure}[ht]
    \centering
    \includegraphics[width=0.9\linewidth]{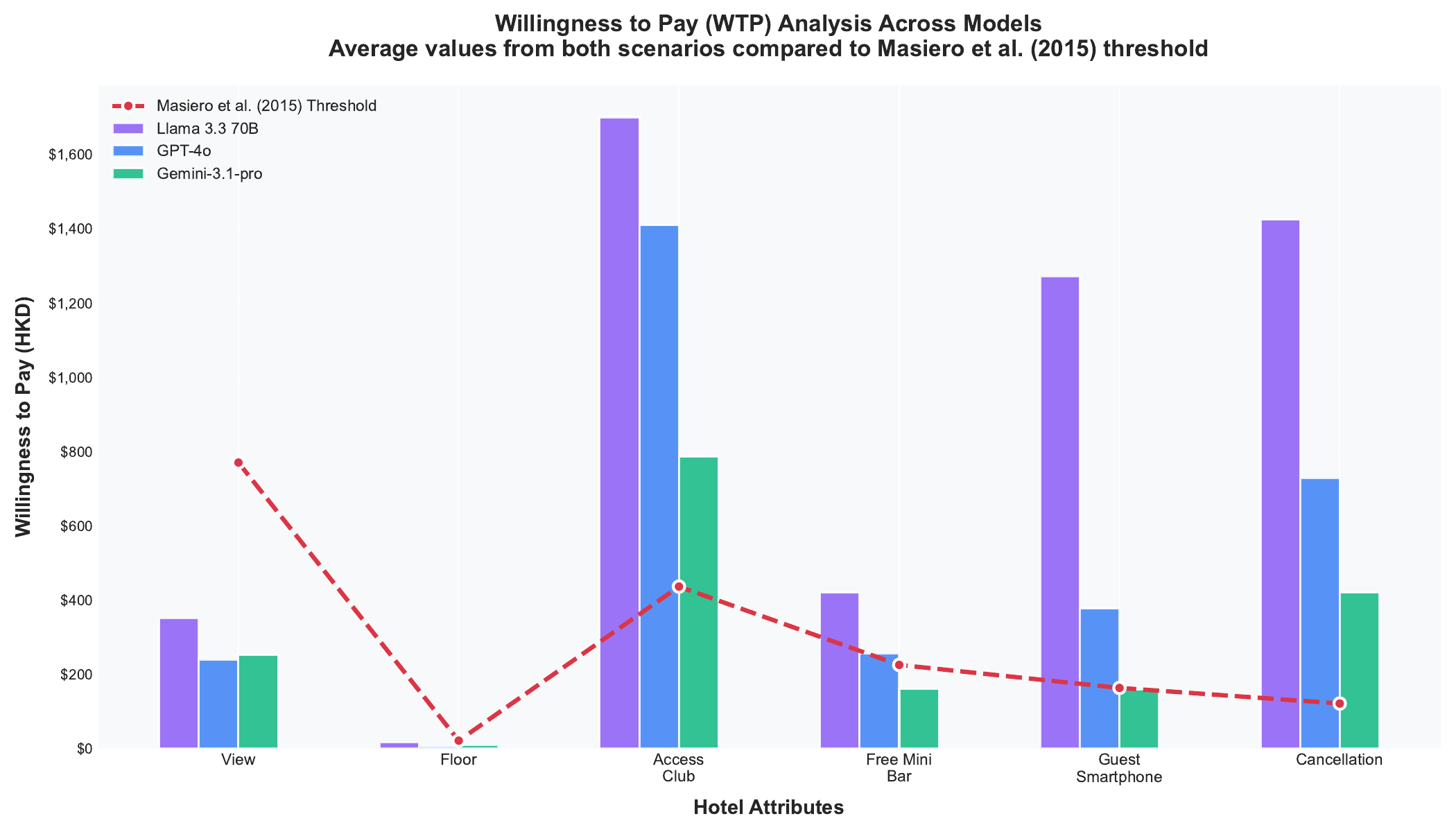}
    \caption{WTP values averaged across the two experimental scenarios for the different models. The dashed red line represents the threshold values from Masiero et al. (2015).}
    \label{fig:wtp_avg_no_info}
\end{figure}

To analyze the final impact on the WTP, we show the WTP scores in Table~\ref{tab:models_no_info_WTP_sec}. This table shows differences in WTP values when comparing scenarios, although similar size orders are shown to Table~\ref{tab:models_no_info_WTP}. Figure~\ref{fig:wtp_avg_no_info} shows the averaged WTP values over both runs. This figure clearly shows how on average all models overestimate the importance of access club, guest and cancellation compared to the human WTP and underestimate the importance of the view compared to our human reference. The importance of the floor is in line with the human importance given to this attribute. For the mini bar, we have a small underestimation from Gemini-3.1-Pro and the smartphone is especially overestimated by Llama~3.3~70B.

\begin{table}[ht]
    \centering
    \scalebox{0.85}{\begin{tabularx}{16cm}{l!{\vrule}lllll}
    \toprule
         & Llama~3.3~70B & GPT-4o & Gemini-3.1-Pro & \cite{MASIERO2015117}\\ \midrule
        view & 192.81  & 364.91 & 240.54 & 771\\
        floor &   27.52  & 3.95 & 10.65 & 22\\
        access club & 1529.29  & 1472.64 & 724.58 & 437\\
        free mini bar &  516.53  & 210.79 & 156.96 & 226 \\
        guest smartphone &  1258.56 & 378.25 & 189.80 & 164 \\
        cancellation &  1234.26  & 746.95 & 403.39 & 122\\
    \bottomrule
    \end{tabularx}}
    \caption{Overview of the WTP for the attribute per model calculated when the order of the alternatives was switched.}
    \label{tab:models_no_info_WTP_sec}
\end{table}

\subsubsection{Prompt Robustness}
We conducted three different types of prompt robustness tests. First of all, we looked at the rephrasing of the scenarios, the different prompts can be found in~\ref{app1}.

\begin{figure}[ht]
    \centering
    \includegraphics[width=0.8\linewidth]{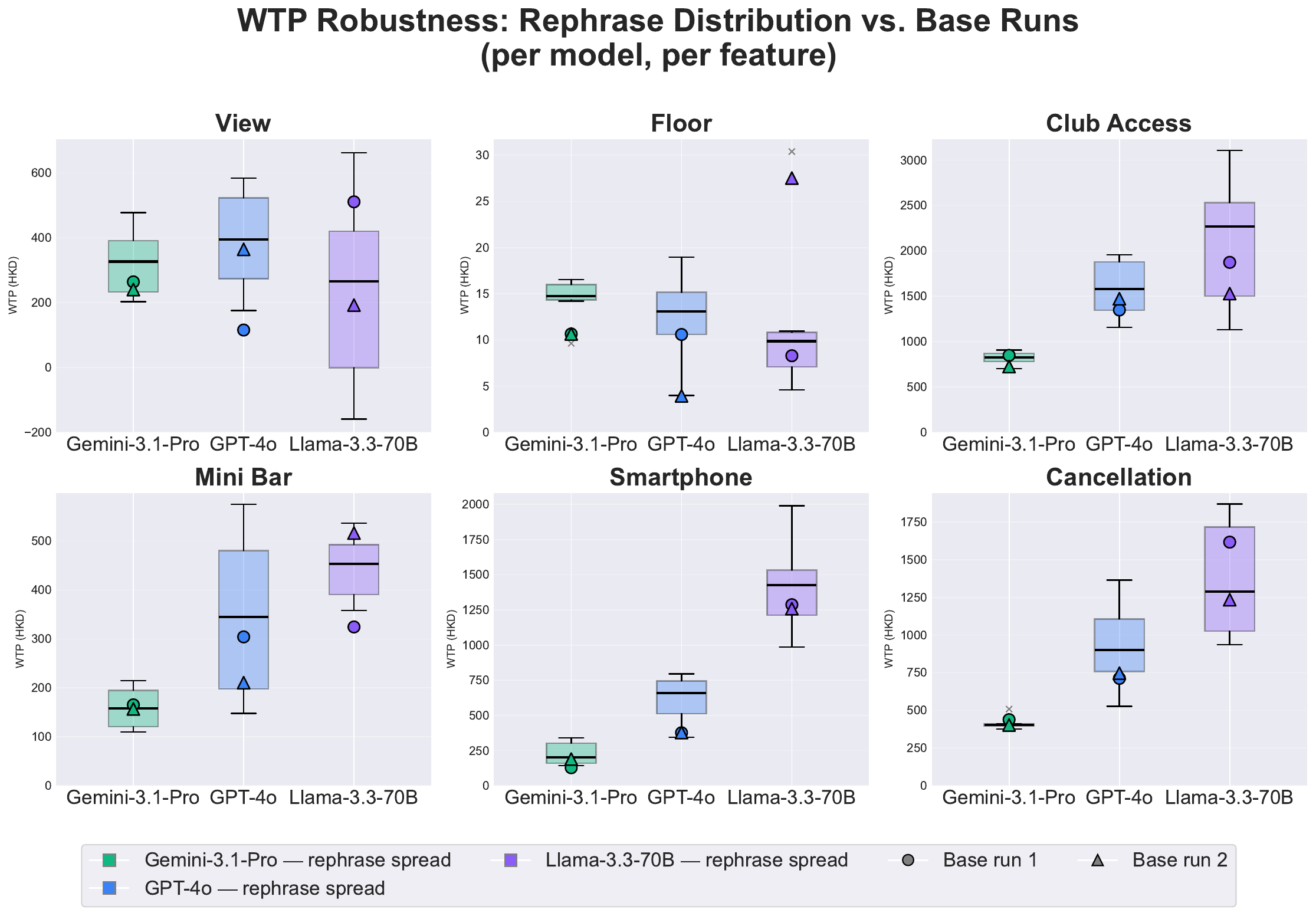}
    \caption{This figure shows boxplots calculated over WTP values derived from the experimental set-up of the 3 different paraphrased results for both orders for the no information case. The baseline results reported throughout the study for the no information case are indicated with a circle and triangle, to show where they appear relative to the paraphrased results.}
    \label{fig:paraphrase}
\end{figure}

As shown in Figure~\ref{fig:paraphrase}, the spread in WTP values across different paraphrased results is largely dependent on the model. While Gemini-3.1-Pro is shown to be rather robust in its responses, a larger spread is observed for GPT-4o. Furthermore, for several attributes, such as view, club access, and smartphone, Llama~3.3~70B shows even a wider spread, with even negative values included in the boxplot estimated for Llama~3.3~70B. In general, however, the results from our initial runs are rather representative for the general WTP values for the other experiments. Note that we did not include the base values in the boxplot estimation, but plotted it on top of the boxplots that were made only with the results from the paraphrased experiments.This figure clearly shows how the sensitivity to prompt rephrases is very model-dependent.

\begin{figure}[ht]
    \centering
    \includegraphics[width=0.8\linewidth]{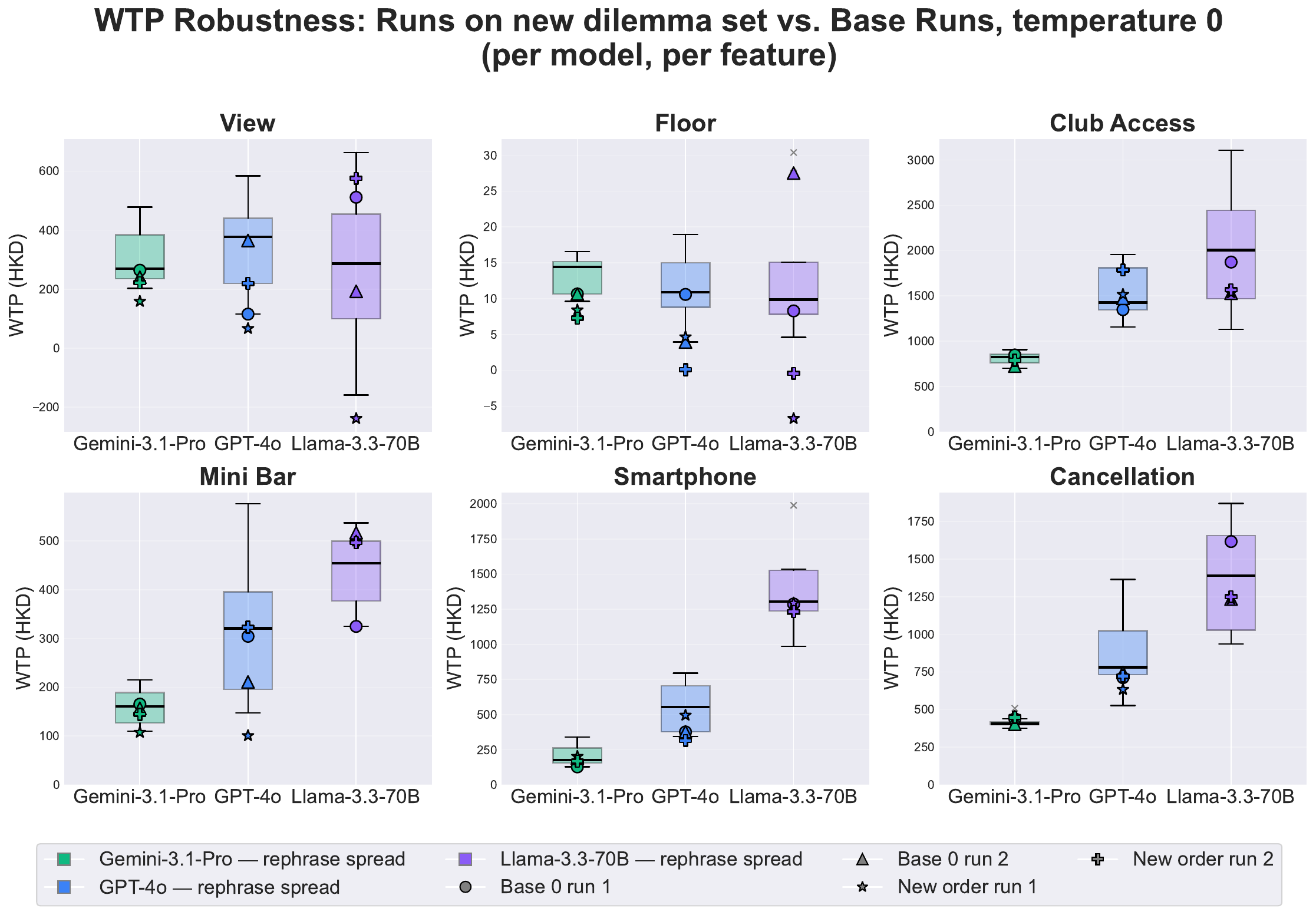}
    \caption{Results of the new dilemma set and old dilemma set plotted against boxplots made from all paraphrased prompts.}
    \label{fig:new_order}
\end{figure}

Next, we included an additional analysis on different generated dilemma combinations. These were now combined with a different random seed (48 instead of 24) to analyze the robustness of our method when using other dilemmas. To assess how these relate to the current results, we use the boxplots from the previous figure, but now trained on both the base settings as well as the rephrased experiments from the previous figure. This allows a thorough comparison on how the WTP values derived with new dilemma settings relate to the ones from the previous dilemmas. This is shown in Figure~\ref{fig:new_order}

The WTP values derived from the new dilemma set for Gemini-3.1-Pro are in general rather similar to the initially derived values. Only for floor, both values fall outside the estimated boxplot. For GPT-4o however, we see that the values are a bit less robust, especially for floor as well, where for the second run, we even found a WTP value close to zero for floor. Similar to our previous experiments, we again see how, in general, LLama provides the least robust results, especially for the view and floor attributes, we find results that lie outside of the boxplots. 

\begin{figure}[ht]
    \centering
    \includegraphics[width=0.9\linewidth]{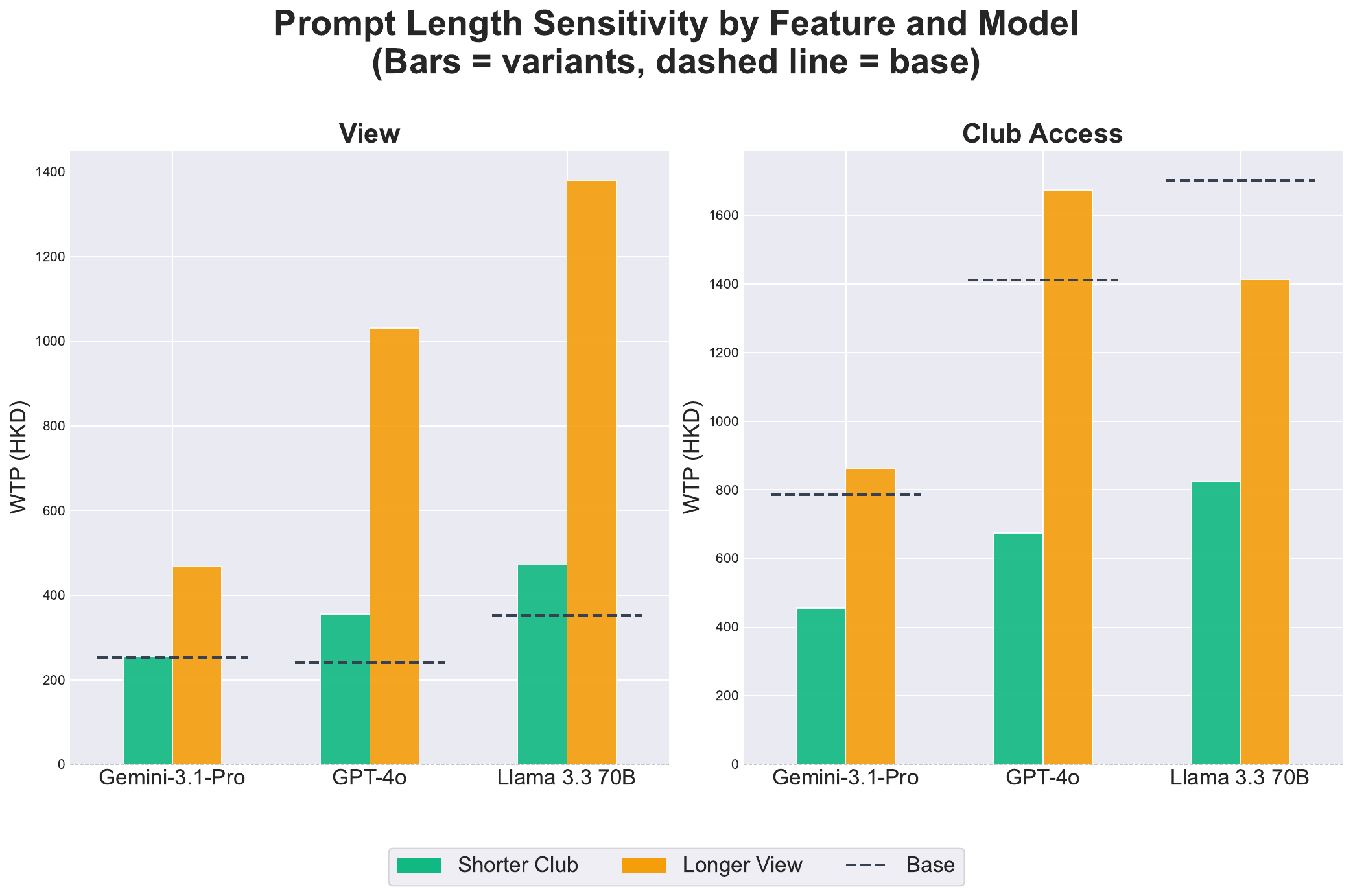}
    \caption{This figure shows the average results over two runs of the different prompt settings: shorter club, longer view, and ou base setting from the paper (no information).}
    \label{fig:prompt_length}
\end{figure}

A final prompt robustness experiment concerns positive feature framing. More specifically, we assess the effect of shortening shortening the explanation of club access, by just mentioning whether or not you have access to the club (shorter club). Additionally, we also investigate the opposite effect, analyzing the effect of including additional information for the view attribute (longer view). More specifically we add the following information to the harbor view: \emph{where beautiful boats lie from very famous people. The view reaches very far allowing the user to see the most beautiful sunsets over the water.} The results of the experiments are summarized in Figure~\ref{fig:prompt_length}.

When looking at the impact of the WTP value for view when making the description of harbour view longer, we see that we can highly influence the WTP values. For Gemini-3.1-Pro, we see that the value almost doubles, and both GPT-4o  and Llama~3.3~70B are even more sensitive, where the averages for two runs are more than four times the initial WTP value averages. For club access, we see that the results are similarly affected when shortening the club access description. The derived average WTP value after this intervention is, for all models, almost half of the initial average WTP value. It is crucial to be aware of this possibility of manipulation of the responses of the LLMs when deploying them in practice. %

\subsubsection{Impact of currencies}
In addition to these experiments using HKD, we also investigated the effect of using USD to assess the effect when converting the prices. The conversion rate used was $1$ HKD $= 0.13$ USD. The final models are made with the initial HKD values, to make them more comparable between the different settings. Additionally, we also analyze whether the currency affects the results gathered from the different LLMs. The models in our current set are all made in the United States. Therefore, we want to check whether there are differences in the estimated WTP when using USD instead of HKD. Note that although we gave the dilemmas to the LLM in USD, we will calculate the WTPs in HKD to make the different results comparable. Thus, while the initial prompts were done in USD, we estimate the multinomial logit model and gather the WTP values using the prices in HKD rather than in USD. 
\begin{figure}[ht]
    \centering
    \includegraphics[width=\linewidth]{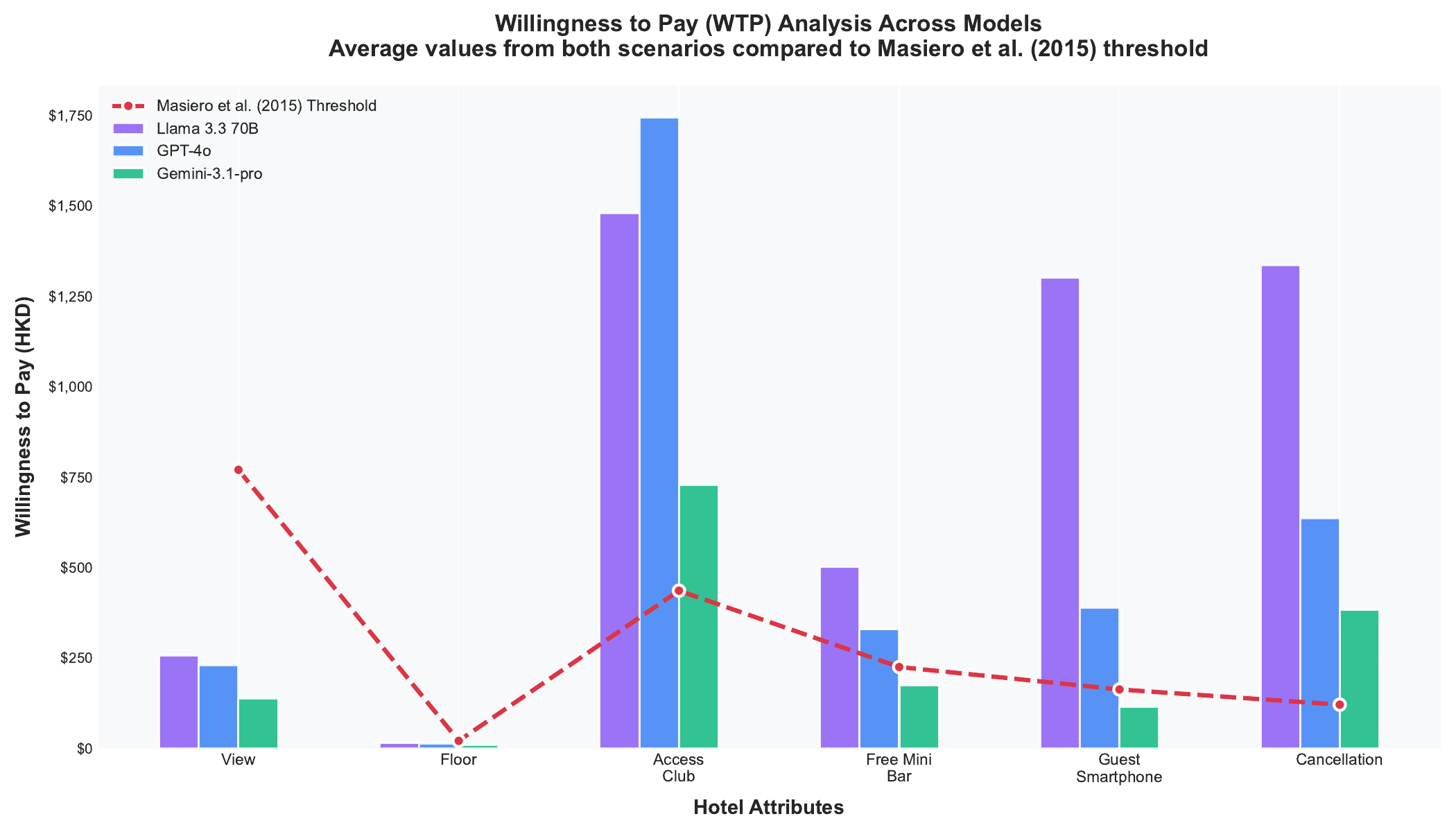}
    \caption{WTP values averaged across the two experimental scenarios for the different models. The dashed red line represents the threshold values from Masiero et al. (2015).}
    \label{fig:wtp_avg_no_info_USD}
\end{figure}

Figure~\ref{fig:wtp_avg_no_info_USD} shows the estimated WTP in HKD when the model was asked the dilemmas in USD. The graph shows comparable results to figure~\ref{fig:wtp_avg_no_info}. However, one of the big differences is the more extreme WTP value of GPT-4o for the access club, compared to the HKD scenario where Llama 70B had the highest WTP value. Additionally, for one of the runs, llama 70B even had a negative value for the WTP for floor, meaning that the model prefers lower floors over higher ones. In general, we find more extreme behavior for some models and features, but less extreme for others (i.e. view). Overall, however, the results are rather in line with one another, showing the models are rather robust against currency changes.

\subsubsection{Robustness to a non-zero temperature setting}
Given that the other experiments were all conducted at a temperature 0, to uncover the learned WTP values of the models, we now also investigate the robustness of these values against a temperature change. In these experiments, we have conducted 10 runs at a 0.7 temperature, a commonly used default that allows for moderate stochastic variation without making responses excessively unstable. 
\begin{figure}[ht]
    \centering
    \includegraphics[width=0.8\linewidth]{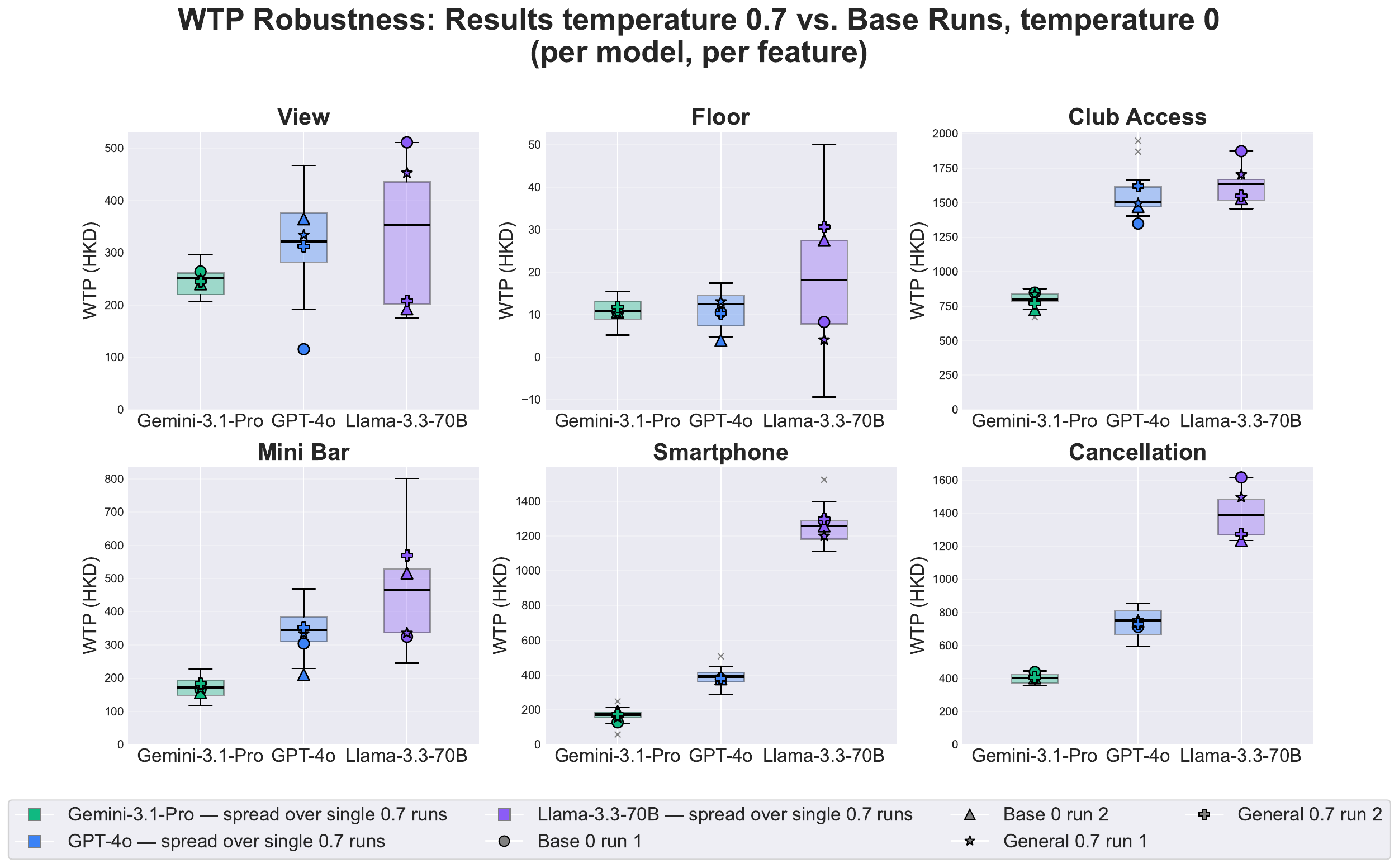}
    \caption{The boxplots are generated from the WTP values derived from the multinomial logit models calculated per run, 10 runs at a temperature of 0.7 for both orders of the dilemmas. The Star and Plus symbol represent the multinomial logit models derived from all results at 0.7 temperature at the first and second run respectively. The Base results plotted, are the WTP values from the initial runs at a zero temperature.}
    \label{fig:temp_robustness}
\end{figure}

Figure~\ref{fig:temp_robustness} plots the WTP values from the temperature-0 run alongside the WTP estimates from the multinomial logit models fitted to all 10 runs at temperature 0.7, reported separately for each dilemma order. The boxplots represent the distribution of WTP values obtained by estimating the model each of the 10 individual runs at temperature 0.7.%

We see how the spread highly depends on the underlying model. Llama~3.3~70B, shows again the largest spread, while Gemini-3.1-Pro shows the most consistent results. Additionally, we see that in general the runs at temperature 0 show representative results for the WTP values at 0.7. Similarly, also the single WTP value derived from all runs provides similar results as our single runs on temperature 0.

\section{Discussion} \label{sec:discussion}
In the next sections, we discuss the overall findings from our experiments, insights into when LLM WTP estimation breaks down, and provide insights into the practical implications of our findings.

\subsection{Overall Findings}
In this section, we will cover the overall findings of the paper structured per hypothesis as introduced in Section~\ref{intro}.

\paragraph{Hypothesis 1: Structural Preference Emergence} Our first set of experiments analyzed whether larger LLMs exhibit interpretable WTP estimates and whether these estimates differ per LLM. By modeling LLM choices using multinomial logit models and interpreting the resulting coefficients through WTP, we obtain a structured and interpretable view of model preferences that goes beyond standard evaluation metrics. Even in the absence of user-specific information, this approach reveals substantial heterogeneity across models in both explanatory power and valuation patterns. Larger models tend to achieve higher $R^2$ values and mostly exhibit economically coherent preferences, assigning positive utility to desirable attributes and negative utility to price. Smaller models, however, frequently exhibit high order bias or yield multinomial logit models with limited explanatory power. However, also larger models can break down, we assess the different scenarios in which breakdowns occur in the following section.  %
Comparing model-derived WTP estimates highlights systematic differences in how LLMs value specific attributes. Across models, certain features are consistently underweighted compared to the human reference. Furthermore, we find large differences between the WTP values for different models. In general, both Llama~3.3~70B and GPT-4o provide larger estimates than Gemini-3.1-Pro. %

\begin{figure}[ht]
    \centering
    \includegraphics[width=0.9\linewidth]{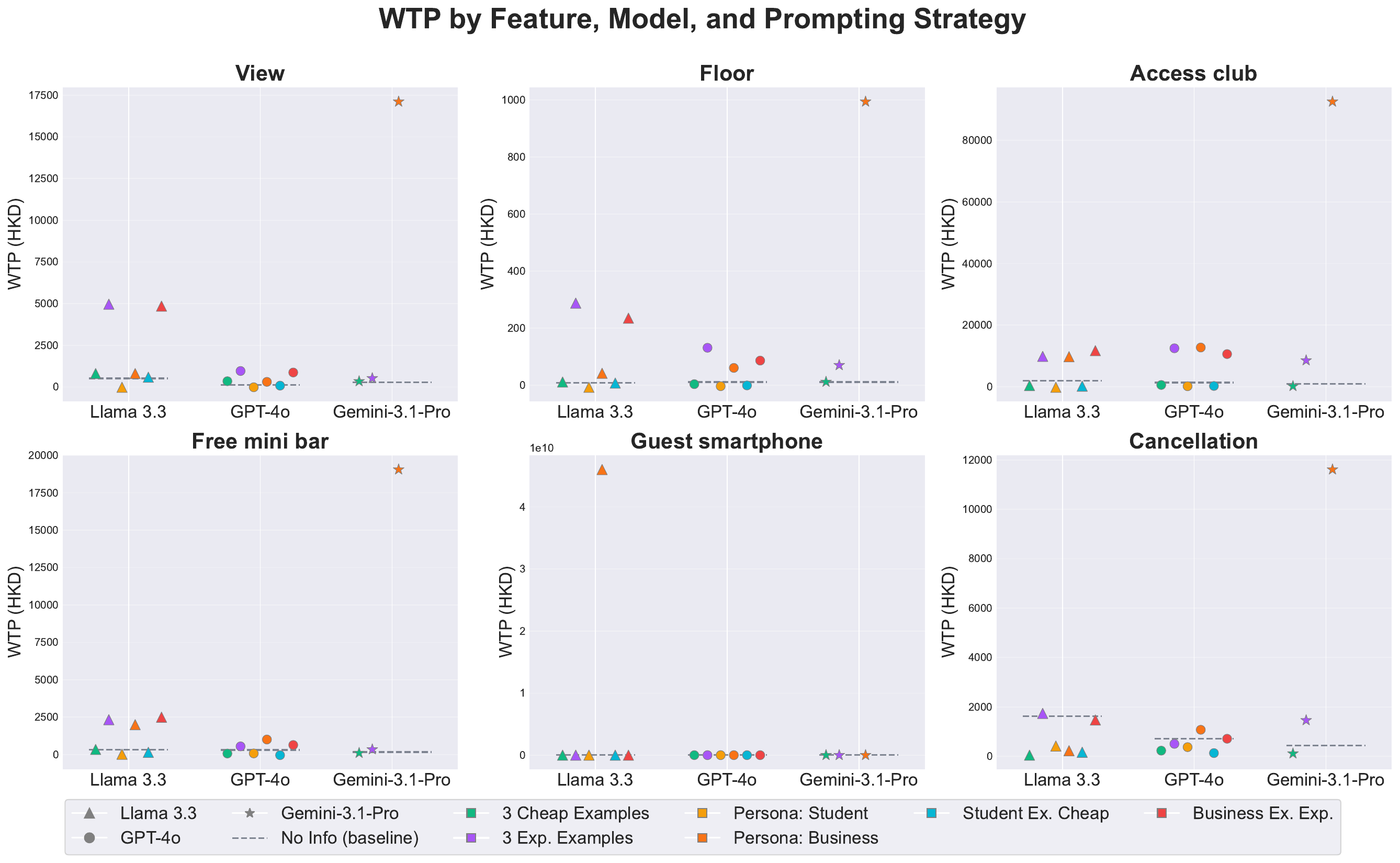}
    \caption{Overall overview WTP values of different prompting strategies compared to the No information baseline.}
    \label{fig:overview}
\end{figure}

\paragraph{Hypothesis 2: Context-driven parameter shifts}
Our second hypothesis covers the question of whether or not including user-specific information can lead to a shift in the estimated WTP values. Figure~\ref{fig:overview} clearly shows that it is possible to shift the WTP values both upward and downward by providing different context information on the user. More specifically, we find that, in general explicitly conditioning models on preferences for more expensive options, whether through in-context learning, persona assignment, or a combination of both, leads to highly amplified WTP estimates. In contrast, conditioning on preferences for cheaper options yield lowered WTP values compared to when no information was added in our experimental setup. Note that, in addition to some feature-specific exceptions, such as the increased WTP for all models for the view attribute when adding three cheap examples, another important exception arises when using persona-based conditioning with Gemini-3.1-Pro. In this case, the persona alone induces similarly extreme behavior, with the model consistently selecting the lowest-priced option regardless of other attributes.

\paragraph{Hypothesis 3: Prompt-dependent shifts in WTP values}
Our final hypothesis looks at how WTP values can be shifted through representation changes, like order switching, prompt perturbations, using another currency, and under different temperature settings. In general, we find that models display variable WTP values depending on the previously mentioned analyses. Switching the order of the dilemmas affects the derived WTP values. Furthermore, paraphrasing also results in a large spread in terms of estimated WTP values. This spread is, however, dependent on the underlying LLM: Gemini-3.1-Pro exhibits the smallest overall spread, followed by GPT-4o. Llama~3.3~70B seems to be least able to produce stable WTP values under paraphrased prompts. This robustness difference across the different models is also shown across the different dilemma set. Furthermore, similar findings are shown when changing the temperature to 0.7. Here we again find more stability in the WTP values estimated for Gemini-3.1-Pro and GPT-4o compared to Llama~3.3~70B. Similar WTP values are found when comparing the general multinomial logit model over all 0.7 runs with the single runs at a zero temperature. Furthermore, we find the effect of positive feature framing within the prompts striking. By removing positive framing from the club access feature, we can half the estimated WTP value for this feature. Additionally, we can increase the estimated WTP value for view by double the amount for Gemini-3.1-Pro and four times for GPT-4o and Llama~3.3~70B by including positive framing for this attribute. The impact of changing the currency is in general not as big as expected, with a few exceptions, such as the more extreme WTP value of GPT-4o for the access club feature.

\subsection{Breakdown cases in LLM WTP estimation}
As shown throughout the paper, we identify several scenarios in which LLM-based WTP estimation breaks down and yields economically invalid or uninterpretable results. These scenarios are not incidental, but instead reveal systematic limitations of the models and prompting strategies used. Given that these breakdown cases deserve explicit attention, we provide an overview of the main scenarios, discussing the underlying model behavior and identifying which models and experimental conditions are affected.

\paragraph{Order Bias} It is well known that models are prone to order bias, preferring a first option over subsequent ones~\citep{allouah2025your}. We observe this behavior in our experiments as well, particularly among smaller models (see Section~\ref{sec:models}), though the degree varies across experiments. For example, we saw that Llama~3.2~3B always preferred the first option, indicating a deterministic order effect. At the same time, Llama 3.3 70B also exhibits order bias, but to a smaller extent, as reflected by the alternative specific constant reported in Table~\ref{tab:models_no_info}. In contrast, GPT‑4o shows little to no evidence of systematic order bias. From a WTP perspective, order bias undermines the interpretation of estimated trade-offs, since choices no longer reflect attribute-level preferences but positional heuristics.

\paragraph{Distinct rule-based decision boundary} When assigning the \textit{Student persona} to Gemini-3.1-Pro (both with and without additional ICL examples), the LLM preferences are perfectly predicted by solely looking at the price. In this case, our WTP set-up is not necessary to understand the LLM behavior. This was steered by the strict following of the assigned persona where it was asked to make the most budget-friendly choice. While this behavior is internally consistent with the assigned persona, it departs from realistic human decision-making, where price alone is rarely the sole determinant. This is particularly the case in trade-offs involving attributes such as cancellation policies, where paying a slightly higher price may reduce expected costs under uncertainty. 

\paragraph{Negative WTP values due to negative price coefficients} In several experimental conditions, we estimate positive price coefficients in the multinomial logit models, implying negative WTP for certain amenities. Such estimates are not economically meaningful and instead constitute a clear failure signal of the WTP values estimated within these models. Negative WTP emerges both for smaller models, i.e. for Llama~3.1~8B when adding three expensive examples with or without the business persona and, more surprisingly, for Gemini‑3‑Pro when combining a business persona with in‑context learning examples emphasizing expensive choices. In this extreme configuration the model appears to equate higher prices directly with higher comfort, strictly following the persona instruction to make the stay \textit{“as comfortable as possible.”}. This induces a monotonic preference for higher-priced alternatives, despite the fact that higher prices are not necessarily associated with greater utility across all attributes.  Consequently, the model fails to construct stable attribute-level trade-offs, leading to economically incoherent utility estimates. We therefore interpret negative WTP values due to negative price coefficients not as substantive preference signals, but as indicators that the WTP extraction framework breaks down under certain prompting conditions.

\subsection{Practical Implications}
Overall, we thus find that larger LLMs can exhibit interpretable WTP values. However, LLMs can be steered towards extreme WTP values when provided with cues that the user might prefer more expensive options. Additionally, positive framing can also highly influence and increase attribute's WTP values. Consequently, caution is warranted when deploying LLMs as travel assistants. Based on our results, we highlight several practical implications for the deployment of LLMs in a travel-agent decision-support setting.  

\paragraph{Transparency in LLM decision-making} Firstly, our experiments demonstrate that the proposed methodology provides a practical way to make LLM decision behavior more transparent. This is highly relevant for real-world deployment, as LLMs are increasingly used in decision-support settings where they can shape economically meaningful outcomes. Greater transparency is therefore essential not only for understanding model behavior but also for auditing whether recommendations are robust, unbiased, and aligned with user preferences.

\paragraph{Implement larger models as travel assistants} Secondly, our findings suggest that smaller models are generally less suitable for acting as decision-makers on behalf of users. In particular, these models exhibit pronounced order bias, systematically favoring the first presented option irrespective of its substantive content. This has important implications for practitioners considering low-cost models for applications involving subjective recommendations. While order bias is substantially reduced for larger models, it does not disappear entirely, as evidenced by the statistically significant alternative-specific constants observed in our analysis. Similar patterns have been reported in related work, i.e., \cite{allouah2025your}, underscoring that order sensitivity remains a relevant concern even for more capable models.

\paragraph{Be aware of shifted WTP behavior depending on the user context}Thirdly, in terms of setting up these models, it is crucial to handle the addition of user information carefully, as different ways of including this information (persona, ICL examples, or both) result in different WTP values. It is also important to understand the different behavior across models, as shown these can lead to large differences in estimated WTP values.

\paragraph{Hedge your assistant against positive framing} Fourthly, it is crucial to be aware of the sensitivity of these LLMs to positive framing. This suggests that small differences in wording can meaningfully steer model outputs, as is also shown in literature, i.e. \citet{zhu2023promptrobust, elazar2021measuring}. Suppliers can be able to exploit this sensitivity by including positive framing in their product or hotel descriptions. Therefore, additional research is required into the best ways of hedging models against this positive framing.

\paragraph{Prefer human-in-the-loop deployment over full autonomy} Finally, when deploying LLMs as travel assistants that act on behalf of users, unconstrained delegation is risky: models can exhibit positional heuristics (order bias), collapse decisions into single‑attribute rules, or infer economically incoherent trade‑offs under certain prompting conditions. These breakdown cases imply that LLM‑based decision agents should not be treated as unconditional preference optimizers. Instead, deployment should include explicit guardrails, such as order randomization, sanity checks on implied trade‑offs (e.g., monotonic price preferences), and fallback mechanisms to human‑readable rules when dominance or incoherent WTP patterns are detected.

Taken together, these findings suggest that practitioners should exercise caution when deploying LLMs for subjective decision-making. %

\section{Conclusion}
Large language models are increasingly deployed in systems that support or automate decision-making on behalf of users. In many such applications, such as travel assistance or purchasing support, models are required to make subjective choices in settings where no objectively correct answer exists. This raises fundamental questions about whether LLMs can meaningfully express preferences, how these preferences might be shifted through user information or prompt formulation, and to what extent users can rely on these systems in practice. 

In this paper, we introduce an economic framework to study LLM decision-making in subjective choice contexts. Focusing on a hotel-based decision setting, we elicit model choices between paired alternatives of hotel rooms and repeat these decision tasks across multiple runs. We analyze the resulting choices using multinomial logit models and derive implied WTP estimates, which we contextualize using a human reference. In addition to a baseline setting, we evaluate model behavior across a range of realistic conditions, including the provision of information about users’ past choices, the use of persona-based descriptions, and combinations of both. This study thus focuses on a hotel-based decision setting, future research could further extend the experiments to include other domains.

Our results show that for larger LLMs meaningful WTP values can generally be derived. Additionally, LLM WTP values can be highly steered upward and downward depending on the user information included. The ranges in which these values can be steered are very wide and therefore, adding this information should be handled with care as to not invoke extreme model behavior. Additionally, we highlight the sensitivity of model outputs to framing and attribute description throughout our experiments. Furthermore, we summarize several breakdown cases of WTP estimation in LLMs: order bias, distinct rule-based decision boundary, and negative WTP values. 

Taken together, these findings form important practical implications. We find that LLM-derived WTP values are dependent on the model and can be further shifted depending on the included user context. Furthermore, LLMs are highly sensitive to positive framing of features, against which LLMs should be hedged. Finally, given the different breackdown cases on which we elaborate throughout the paper, a human-in-the-loop deployment of a travel assistant is more safe than when providing the LLM with full autonomy. In future work, we would like to investigate the inclusion of more nuanced persona representations to see if they mitigate the extreme behavior witnessed in the results of Gemini-3.1-Pro. Furthermore, a comparison study to a recent human benchmark would provide additional insights into human-LLM preference alignment. Additionally, note that the WTP values derived in this paper are average marginal WTP values, rather than explicit switching thresholds. Complementary to our research additional analysis into the explicit switching thresholds, as conducted in~\cite{cedro2025cash,FULMAN2025104542} includes an interesting direction for future research. A final important avenue for further work is the explicit integration of LLMs into a travel-agent pipeline, to assess whether our findings fully generalize to real-world agentic settings.

\section{Acknowledgments}
We acknowledge the support of the Research Foundation Flanders (FWO), Grant G0G2721N. Additionally, we thank the Antwerp Center on Responsible AI (ACRAI) for their support.

\section{Declaration of generative AI usage}
During the preparation of this work, the authors used ChatGPT for minor text editing purposes. After using this tool, the authors reviewed and edited the content and take full responsibility for the content of the published article.
\appendix
\section{Paraphrased Alternatives for robustness analysis}
\label{app1}
In this appendix, the three paraphrased alternatives are shown that are used for the robustness analysis.
Throughout the robustness analyses, we tested paraphrased versions for the scenarios only, the main prompt structure remained the same.

\begin{tcolorbox}[scenariopromptbox]
\texttt{A room on floor \textit{{{floor}}}has been reserved, offering a view of the \textit{{{view}}}.\\
The guest \textit{{{club\_access}}} granted access to the hotel club, offering exclusive amenities including breakfast and evening cocktails in the panoramic restaurant. \\
The room features a complimentary mini bar stocked with \textit{{{free\_mini\_bar}}} and a guest smartphone is \textit{{{guest\_smartphone}}}\\ 
The reservation is \textit{{{cancellation}}} at a nightly rate of \textit{{{price\_per\_night}}} per night.}
\end{tcolorbox}

where club\_access  refers to \textit{is} or \textit{is not}.

The second paraphrasing is the following: 
\begin{tcolorbox}[scenariopromptbox]
\texttt{Room on floor \textit{{{floor}}}, \textit{{{view}}} view.\\
Club access \textit{{{club\_access}}} breakfast and evening cocktails served at the panoramic restaurant. \\
Mini bar with \textit{{{free\_mini\_bar}}} and smartphone \textit{{{guest\_smartphone}}}\\ 
Cancellation policy: \textit{{{cancellation}}}. Price: \textit{{{price\_per\_night}}}/night.}
\end{tcolorbox}

where club\_access  refers to \textit{included:} or \textit{not included: no }.

The third paraphrasing is the following: 
\begin{tcolorbox}[scenariopromptbox]
\texttt{The room is located on the \textit{{{floor}}} floor and overlooks the \textit{{{view}}}.\\
You \textit{{{club\_access}}} have access to the hotel club, where you can enjoy perks like breakfast and evening cocktails at the panoramic restaurant. \\
You'll enjoy a free mini bar with \textit{{{free\_mini\_bar}}}and a smartphone\textit{{{guest\_smartphone}}}.\\ 
Your booking is \textit{{{cancellation}}}, priced at \textit{{{price\_per\_night}}} per night.}
\end{tcolorbox}

where club\_access  refers to \textit{'ll} or \textit{won't}.

\section{Adjusted WTP scores Masiero et al. (2015)}
We have included the unadjusted WTP values as calculated in~\cite{MASIERO2015117} throughout the paper. However, these values should also be adjusted for inflation to today. Therefore, we have also calculated the adjusted WTP values by dividing these values by the Consumer Price Index (CPI) of 2015 and multiplying it by the CPI of 2025. We used the following values: for the CPI of 2015 92.8 and for the CPI of 2025 109.4~\footnote{https://tradingeconomics.com/hong-kong/consumer-price-index-cpi}. These calculations, however, have a minor effect and do not cause conflicting findings to the ones mentioned above. Table~\ref{tab:adjusted_wtp} shows the original and adjusted WTP values.

\begin{table}[ht]
    \centering
    \begin{tabular}{l|ll}
    \toprule
     & \makecell{\cite{MASIERO2015117} \\} & \makecell{\cite{MASIERO2015117} \\ — CPI-adjusted to 2025} \\ \midrule
        view &  771 & 908.92\\
        floor & 22 & 25.94 \\
        access club & 437 & 515.17 \\
        free mini bar & 226 & 266.43\\
        guest smartphone & 164 & 193.34 \\ 
        cancellation & 122 & 143.82\\
         \bottomrule
    \end{tabular}
    \caption{Adjusted WTP values from 2015 to 2025 as shown in~\cite{MASIERO2015117}}
    \label{tab:adjusted_wtp}
\end{table}

\bibliography{my_bibliography}

\end{document}